\documentclass[conference]{IEEEtran}
\usepackage{times}

\usepackage[numbers]{natbib}
\usepackage{multicol}
\usepackage[bookmarks=true]{hyperref}
\usepackage{graphicx}
\usepackage{cuted}
\usepackage{caption}
\usepackage{siunitx}
\usepackage{amsmath} 
\usepackage{amssymb}  
\usepackage{diagbox} 
\usepackage[noabbrev,capitalise]{cleveref}
\usepackage[table, dvipsnames]{xcolor}%
\usepackage{bbm}
\usepackage{booktabs}
\usepackage{macros}
\usepackage{subcaption}
\usepackage{pgfplots}
\usepackage{pgfplotstable}
\usepackage{tikz}
\pgfplotsset{compat=1.18} 
\usepackage[dvipsnames]{xcolor}
\usepackage{arydshln}
\definecolor{IdealColor}{HTML}{457B9D}
\definecolor{CEMColor}{HTML}{E07A5F}
\definecolor{DRColor}{HTML}{D4B886}
\definecolor{ROAColor}{HTML}{2F4858}
\definecolor{ActuatorNetColor}{HTML}{52936C}
\definecolor{UANColor}{HTML}{C45552}
\definecolor{Sage}{HTML}{94A89A}
\definecolor{DustyRose}{HTML}{D4A5A5}
\definecolor{SlateBlue}{HTML}{9A97C4}
\definecolor{WarmTaupe}{HTML}{B6A392}

\begin{document}

\title{
Bridging the Sim-to-Real Gap for
Athletic Loco-Manipulation}

\author{\authorblockN{Nolan Fey, Gabriel B. Margolis, Martin Peticco, and Pulkit Agrawal}
\authorblockA{Improbable AI Lab\\
Massachusetts Institute of Technology, Cambridge, MA 02139}
}

\maketitle

\footnotetext[1]{Authors are also affiliated with Computer Science and Artificial Laboratory (CSAIL), the Laboratory for Information and Decision Systems (LIDS), and the MIT-IBM Watson AI Lab at MIT. Correspondence to {\href{mailto:nolanfey@mit.edu}{\texttt{nolanfey@mit.edu}}}}

\begin{abstract}
Achieving athletic loco‑manipulation on robots requires moving beyond traditional tracking rewards—which simply guide the robot along a reference trajectory—to task rewards that drive truly dynamic, goal-oriented behaviors. Commands such as “throw the ball as far as you can” or “lift the weight as quickly as possible” compel the robot to exhibit the agility and power inherent in athletic performance. However, training solely with task rewards introduces two major challenges: these rewards are prone to exploitation (reward hacking), and the exploration process can lack sufficient direction. To address these issues, we propose a two‑stage training pipeline. 
First, we introduce the Unsupervised Actuator Net (UAN), which leverages real‑world data to 
bridge the sim-to-real gap for complex actuation mechanisms without requiring access to torque sensing.
UAN 
mitigates reward hacking by ensuring that the learned behaviors remain robust and transferable. 
Second, we use a pre‑training and fine‑tuning strategy that leverages reference trajectories as initial hints to guide exploration. 
With these innovations, our robot athlete learns to lift, throw, and drag with remarkable fidelity from simulation to reality.
\end{abstract}

\IEEEpeerreviewmaketitle

\section{Introduction}
\label{sec:introduction}

\begin{figure}
   \centering
   \begin{minipage}{\linewidth}
    \centering
    \includegraphics[width=0.9\linewidth]{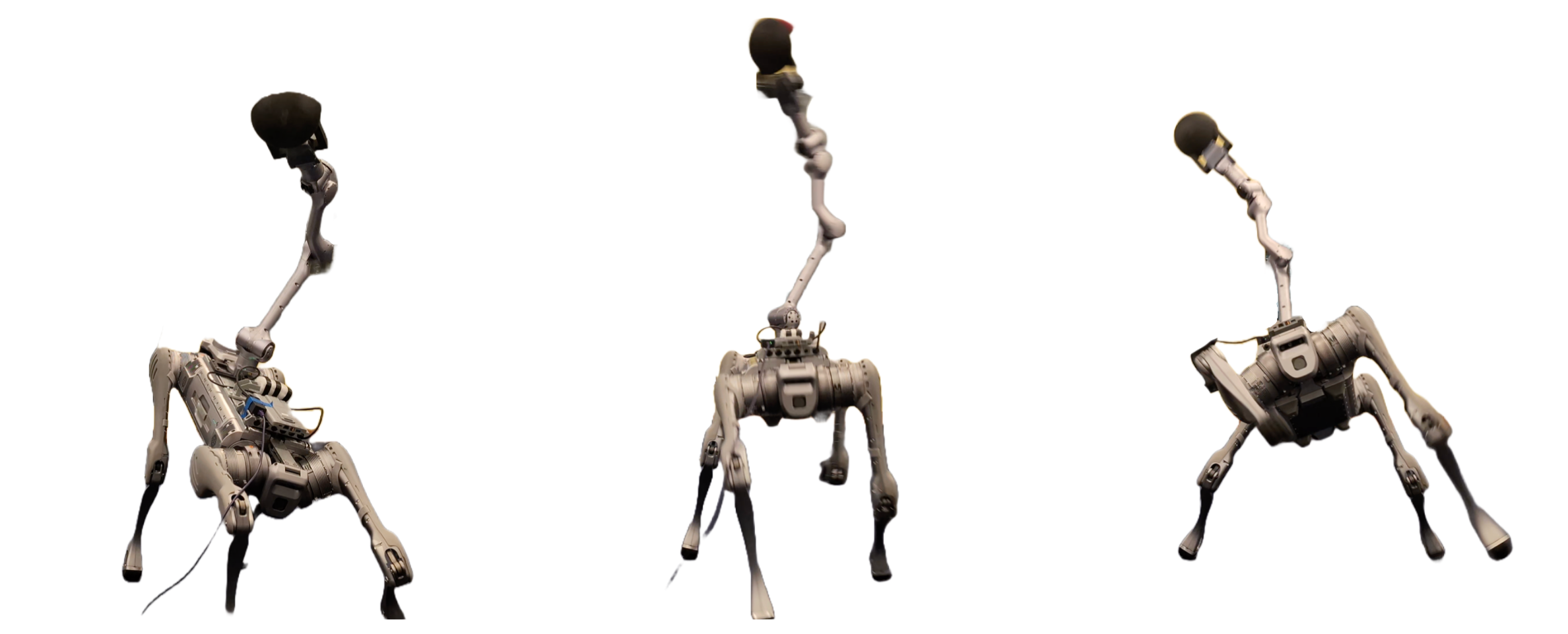}
    
    \textit{Ball Throw}
  \end{minipage}
  
  \vspace{1em}
  
  \begin{minipage}{\linewidth}
    \centering
    \vspace{-0.7cm}
    \includegraphics[width=0.9\linewidth]{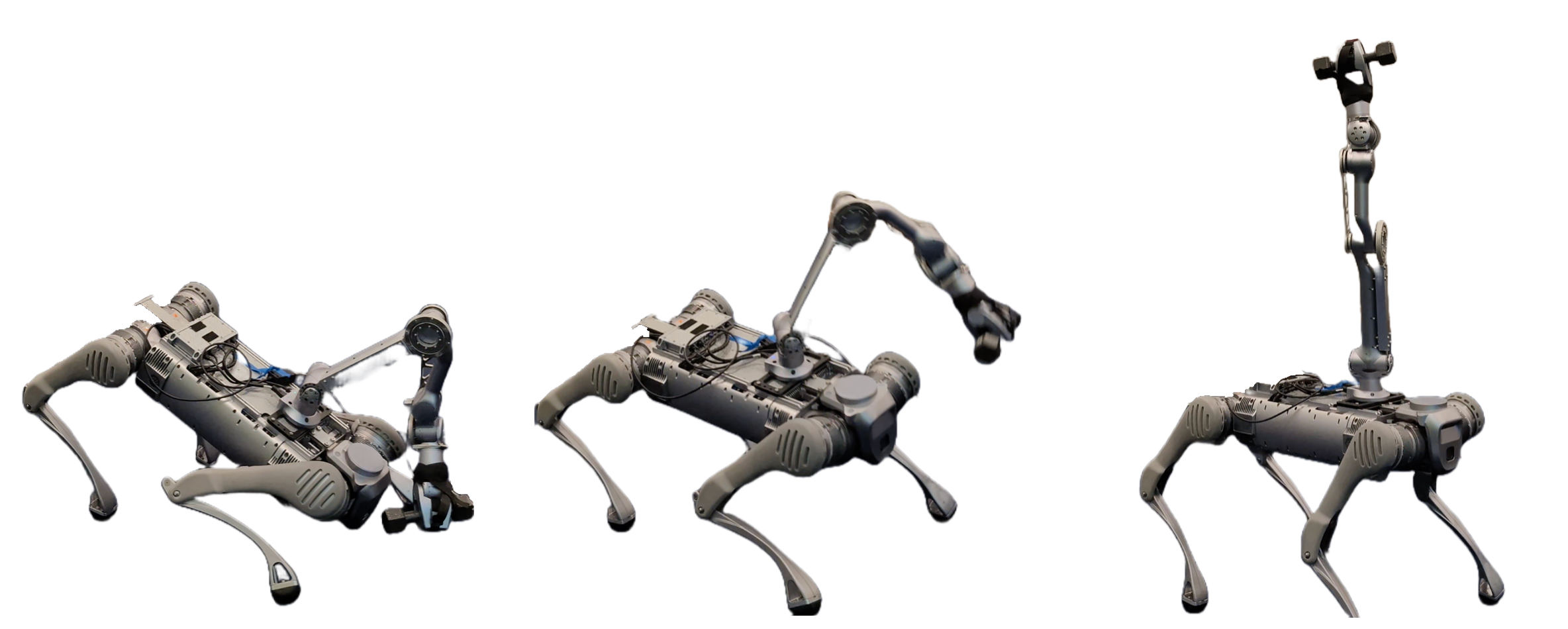}
    
    \textit{Dumbbell Snatch}
  \end{minipage} 
  
  \vspace{1em}
  
  \centering
  \begin{minipage}{\linewidth}
    \centering
    \vspace{0.2cm}
    \includegraphics[width=0.9\linewidth]{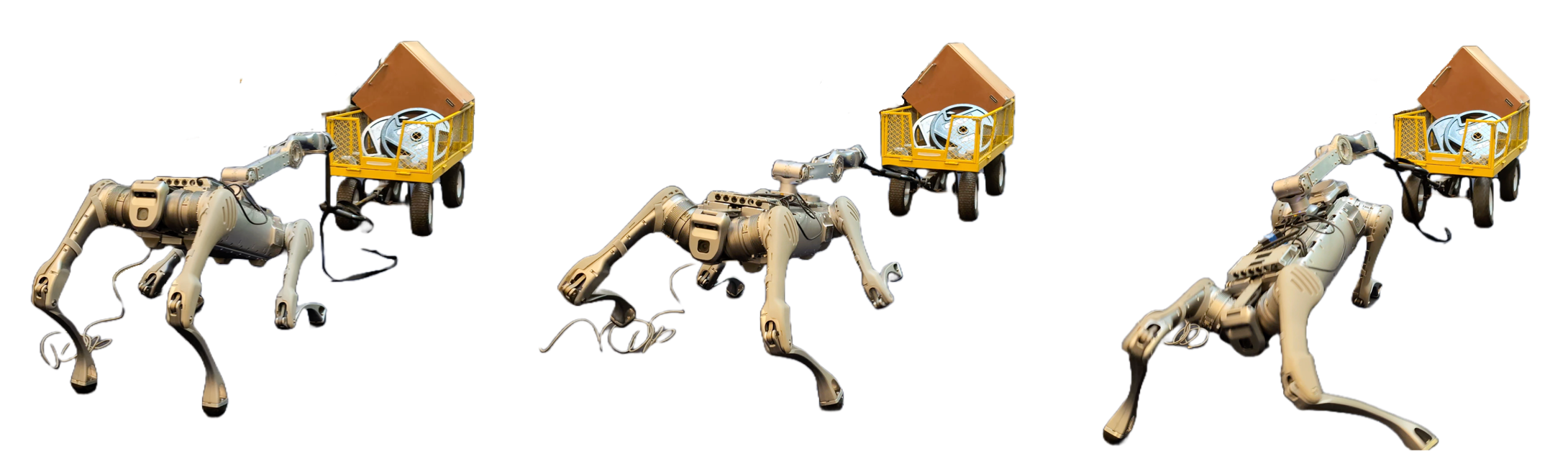}
    
    \textit{Sled Pull}
  \end{minipage}
  \vspace{0.5em} 
  
   \captionof{figure}{\textbf{Sim-to-real transfer of athletic loco-manipulation.} We reduce the sim-to-real gap for a quadruped manipulator by \textbf{learning a corrective model for the simulated actuator dynamics based on real-world data}, formulated as an \textit{unsupervised actuator net (UAN)}. Policies trained with the corrected simulator exhibit improved sim-to-real transfer and push the limits of the robot's physical capabilities in athletic tasks involving whole-body coordination. Videos of the robot's behaviors are available at \url{https://uan.csail.mit.edu/}. }
   \label{fig:throw-comparison}
\end{figure}

General whole-body control comes naturally to animals after years of evolution, yet it remains a long-standing challenge in robotics.
Fluid whole-body motion requires balancing multiple competing tasks and constraints that depend on both the robot's morphology and its environment~\cite{HWBC-Sentis}. Recent work~\cite{fu2023deep, portela2024learning} demonstrates that sim-to-real reinforcement learning (RL), using methods such as Proximal Policy Optimization (PPO)~\cite{schulman2017proximal}, is a promising paradigm for learning these behaviors by leveraging parallel simulations~\cite{rudin2022learningwalkminutesusing}.

For dynamic, goal-oriented loco-manipulation, it is natural to train robots with task rewards—commands like “throw the ball as far as possible” or “lift the weight as quickly as possible” that drive athletic behaviors. However, these task rewards pose two major challenges: (i) they are prone to reward hacking, where the policy exploits imperfections in the simulation, and (ii) the exploration process can lack sufficient guidance. To circumvent these issues, many works on sim-to-real transfer instead train whole-body controllers (WBCs) to track dense reference motions~\cite{cheng2024expressive, fu2023deep, he2024learning, liu2024visual, portela2024learning}. Dense tracking objectives provide strong regularization by constraining the policy to adhere to a reference trajectory—thereby reducing reward hacking—and they offer a structured path for exploration. 
However, this strategy relies on defining high-quality reference commands a priori, which in turn demands access to high-quality reference data. For robots with non-human morphologies like legged manipulators, obtaining such data is particularly challenging, and the resulting reference commands may not capture the optimal, athletic strategies that a policy might otherwise discover.

To fully harness the benefits of task rewards, it is crucial to ensure that the simulation faithfully replicates real-world dynamics. Inaccurate simulation models allow policies to exploit imperfections, leading to reward hacking, particularly so when the reward is underspecified.  Although techniques like domain randomization~\cite{tan2018simtoreal, tobin2017domain, xie2021dynamics} and online system identification~\cite{concurrent_se, kumar2021rmarapidmotoradaptation, long2024hybridinternalmodellearning, margolis2022rapidlocomotionreinforcementlearning, nahrendra2023dreamwaq, radosavovic2023realworldhumanoidlocomotionreinforcement} address this by sampling over parameter distributions, they rely on a priori assumptions that may not fully capture the complex dynamics of real hardware. For instance, harmonic drive actuators exhibit non-linear friction, hysteresis, and lag—behaviors that render traditional proxies like motor current unreliable for torque estimation.

A promising alternative is to enhance the simulation’s physics model directly with real-world data, focusing on accurately modeling the actuator dynamics. With this motivation, we introduce the \emph{Unsupervised Actuator Net} (UAN), a framework for learning corrective actuator models without the need for torque sensors. UAN is trained using reinforcement learning to predict corrective torques, $\delta \boldsymbol{\tau} = \pi_{\rm UAN}(\mathbf{e})$, by minimizing discrepancies between simulated and real-world joint encoder measurements. In doing so, UAN effectively bridges the sim-to-real gap even for robots with complex transmission mechanisms and noisy or unavailable torque measurements.

Building on this enhanced simulation environment, we address the challenge of guided exploration for athletic behaviors. Rather than enforcing strict adherence to a reference trajectory, we propose treating it as a \emph{hint} to guide exploration. In our approach, a WBC is first pre-trained on random base velocities and end-effector pose commands to establish a strong motion prior. Then, to learn a new athletic behavior, we initialize the controller with a reference trajectory and fine-tune it using a task-specific reward—allowing the policy to depart from the reference when beneficial.

In summary, our paper presents an \emph{easy-to-use} training pipeline for whole-body athletic behaviors that reliably transfer to reality. First, we employ the Unsupervised Actuator Net (UAN) to calibrate actuator dynamics and mitigate reward hacking, ensuring our simulator accurately reflects real-world physics. With this improved simulation environment, we then pre-train a whole-body controller (WBC) to establish fundamental motion skills and fine-tune it with task-specific rewards—using a reference trajectory merely as a hint to guide exploration. This integrated approach enables our robot to perform dynamic tasks such as throwing, lifting, and dragging with remarkable fidelity.

\begin{figure*}[htbp]
    \centering
    \includegraphics[width=\linewidth]{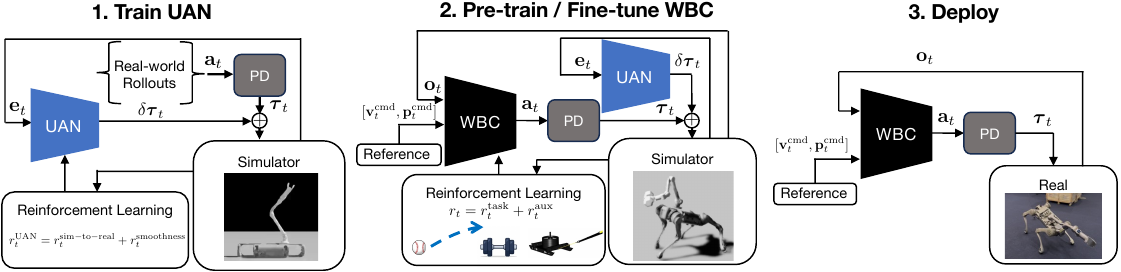}
    \caption{\textbf{Unsupervised Actuator Network (UAN) approach for real-to-sim-to-real.} Our training pipeline involves three steps: 1) Train a UAN to close the sim-to-real gap for actuators with complex transmission mechanisms by mapping a history of joint position and velocity errors, $\mathbf{e}_t$, to corrective torques, $\delta \boldsymbol{\tau}_t$, 2) Pre-train a WBC using random motion references (base velocity and EE pose), then and fine-tune it on an athletic task reward with the UAN in loop, and 3) Deploy. 
    During the fine-tuning phase, the WBC initially tracks the  task-specific reference, and then gradually learns to depart from the reference to maximize task performance.
    }
    \label{fig:TrainDeployDiagram}
\end{figure*}

\section{Method}
\label{sec:method}

Our training pipeline (see Figure \ref{fig:TrainDeployDiagram}) is separated into two phases: 1) real-to-sim calibration (Section \ref{sec:residual_actuator_net}) and 2) WBC training (Sections \ref{sec:pre-train} and \ref{sec:fine-tune}). The real-to-sim calibration phase involves collecting data on the real robot and training a UAN to close the sim-to-real gap for non-ideal actuation mechanisms.
Similar to past work \cite{fu2024humanplushumanoidshadowingimitation, liu2024visual, pan2024roboduet, sferrazza2024humanoidbenchsimulatedhumanoidbenchmark}, our WBC training is split into two distinct sub-phases: pre-training (Section \ref{sec:pre-train}) and fine-tuning (Section \ref{sec:fine-tune}). After pre-training, the policy can track reference trajectories if provided as a sequence of base velocity and end effector pose commands. During the fine-tuning phase, the policy observes a reference task trajectory. This helps warm start exploration when learning a new task because the policy can simply track these commands to achieve reasonable task performance. Through training with the task reward itself rather than a tracking reward, the policy learns how to depart from the reference trajectory to achieve higher task performance. Our simulation environments for the pre-training and fine-tuning phases rely on the same strategies for sim-to-real transfer, including domain randomization (Section \ref{sec:pre-train}) and a UAN (Section \ref{sec:residual_actuator_net}).

Our experiments consider a Unitree B2 quadruped with a modified Unitree Z1 Pro arm mounted on its back. The quadruped is \SI{65}{\centi\meter} tall when standing and weighs \SI{60}{\kilo\gram}, while the arm is \SI{74}{\centi\meter} fully extended and weighs \SI{6.8}{\kilo\gram}. The system has $19$ actuated joints: $3$ for each leg, $6$ for the arm, and $1$ for the gripper. 

\subsection{Unsupervised Actuator Net}

\label{sec:residual_actuator_net}
Some actuators are challenging to model in simulation, especially when they have complex transmission mechanisms. In such cases, standard domain randomization and online system identification techniques may be insufficient, and instead, it is preferable to learn to model the actuator directly from hardware data. Previous approaches rely on output torque sensing \cite{Hwangbo_2019}, which is still uncommon in consumer hardware, to learn how to predict the motor's torque. Alternatively, we propose a method for matching the transition dynamics of the actuator such that
\begin{equation}
    \min_{f_{\rm sim}} ||f_{\rm real}(\mathbf{s}, \boldsymbol{\tau}) - f_{\rm sim}(\mathbf{s}, \boldsymbol{\tau})||.
\end{equation} 
To influence the simulator dynamics, $f_{\rm sim}$, 
we learn a residual model, $\pi_{\rm UAN}(\mathbf{e}$,
that observes a 
history of position and velocity errors,   
$\mathbf{e}$,
and outputs a corrective torque, $\delta \boldsymbol{\tau}$, for the simulator such that
\begin{equation}
    \min_{\pi_{\rm UAN}} ||f_{\rm real} \big(\mathbf{s}, \boldsymbol{\tau} \big) - f_{\rm sim} \left(\mathbf{s}, \boldsymbol{\tau} + \pi_{\rm UAN}\left(\mathbf{e}
    \right) \right)||.
\end{equation}
The corrective torques needed to minimize the transition error
are 
unlabeled, so we parametrize $\pi_{\rm UAN}$ as a neural network and train it with RL. 

\subsubsection{Architecture and observation space} 
The network is designed 
as a 2-layer MLP with layer sizes [128, 128] and ELU activations. It is executed at every simulation time step (5 ms). Assuming each arm joint is identical, a single UAN is shared across all of the arm's actuators, with each actuator being processed independently by the shared network \cite{Hwangbo_2019}. We constrain the observation space to include a history of the past 20 (equivalent to 100 ms) 
position and velocity errors for each relevant actuator. These design choices help prevent overfitting to other aspects of the training data, such as inertial coupling. 
Also, sharing the data across actuators improves data efficiency. For example, the actuator net is trained on data with various loads, as actuators closer to the robot's base generally experience more load than those near the gripper.

\subsubsection{Data collection} We collect data on the real hardware to construct a dataset of transitions $\{ \left(\mathbf{s}_t, \boldsymbol{\tau}_t, \mathbf{s}_{t+1}\right)_i \}_{i=0}^N$ from each actuator. Our intention during data collection was to sufficiently cover the state space to avoid overfitting. Thus, we opted not to use policy data, and instead, collected data with three types of action sequences: 1) square waves, 2) sine waves, and 3) gaussian noise. For the square and sine wave data, we passed torque commands to one actuator at a time, while keeping the rest of the actuators at a fixed position target. We swept $12$ different combinations of amplitude and frequency for each wave, resulting in about $50$ seconds of data for each actuator. For the gaussian noise data, we passed torque commands to all the robot's joints simultaneously. We sampled a new action from a gaussian distribution every $5$ to \SI{400}{\milli\second} for about $5$ minutes. 

\subsubsection{Training Environment}
\label{sec:uan_train_env}
We designed the training environment in Isaac Sim \cite{isaacsim2022} with 4096 parallel environments. We train policies with the RSL-RL implementation \cite{rudin2022learningwalkminutesusing} of PPO \cite{schulman2017proximal} with default hyperparameters, minus a few modifications (see Appendix \ref{sec:appendix_training_details} for the full list of learning algorithm hyperparameters). Following Radosavovic et al. \cite{radosavovic2024learninghumanoidlocomotionchallenging}, we apply a separate, fixed learning rate to the critic while using an adaptive learning rate for the actor.
Additionally, we divide the data of each epoch into four mini-batches for the actor while using the entire batch for the critic, as we found that larger batch sizes produce more stable gradients and result in lower value function loss. 

\subsubsection{Task Design}
For each environment at each timestep, we uniformly sample a real-world transition, $(\mathbf{s}_t, \boldsymbol{\tau}_t, \mathbf{s}_{t+1})_k$, and set the state of the simulator to match $\mathbf{s}_t$ and the initial torque to $\boldsymbol{\tau}_t$. After policy inference, we modify the torque by adding the correction, $\delta \boldsymbol{\tau}_t$, and then step the simulator. We then compute the reward as 
\begin{equation}
    r^{\rm UAN}_t = r^{\rm sim-to-real}_t + r^{\rm smoothness}_t
\end{equation}
where $r^{\rm sim-to-real}_t$ aims to minimize the difference between the real joint position and the simulated joint position, and $r^{\rm smoothness}_t$ biases exploration to gradual deviations. 
For a complete list of reward terms, please refer to Appendix \ref{sec:appendix_training_details}.

Each training episode consists of a \SI{20}{\second} rollout executing the 
torque sequence from the hardware data from 
$\delta \boldsymbol{\tau}_{t}$ to $\delta \boldsymbol{\tau}_{t+20 \SI{}{\second}}$.
Through training on rollouts, the actuator net learns to remain stable across many simulation time steps.

\subsection{Whole-body Controller Pre-training} \label{sec:pre-train}

Before training on task-specific behaviors, we pre-train the WBC to learn foundational trajectory-tracking skills.
Our training scheme builds upon the method proposed in \cite{fu2023deep} by incorporating a strategy for learning to track an EE orientation command.  
As in Section \ref{sec:uan_train_env}, we designed the training environment in Isaac Sim with 4096 parallel environments and trained the policies with PPO \cite{rudin2022learningwalkminutesusing, schulman2017proximal} (using separate learning rates and batch sizes for the actor and critic). 

\subsubsection{Policy Architecture}
The WBC is a control policy, $\mathbf{a}_t = \pi_\theta(\mathbf{o}_{t-H:t})$, where the action at time $t$, $\mathbf{a}_t$, is a vector of position targets for each of the robot's joints and $\mathbf{o}_{t-H:t}$ is an observation history of length $H=10$ timesteps (\SI{200}{\milli\second}). We parameterize $\pi_\theta$ as a $3$-layer multi-layer perceptron (MLP) with layer sizes $[512, 512, 512]$ and ELU activations. The value function approximator network has the same architecture but does not share weights with the policy.

\subsubsection{Observation Space}
The policy's observation space consists of proprioceptive readings from the robot's onboard sensors, including the gravity vector projected in the robot's body frame $\mathbf{g}$,  
a base velocity command $\mathbf{v}_t^{\rm cmd}$, an end effector pose command $\mathbf{p}_t^{\rm cmd}$, the joint positions $\mathbf{q}$, the joint velocities $\dot{\mathbf{q}}$, the previous actions $\mathbf{a}_{t-1}$, and a timing variable $\omega_t = \sin(2 \pi f t)$ with $f=2.2$ Hz corresponding to the gait cycle frequency. 
Additionally, the observation includes a $d$-dimensional task embedding vector $\mathbf{z}_t$ (set to zero during pre-training).

\subsubsection{Sim-to-Real Considerations} Our approach for bridging the sim-to-real gap uses a combination of domain randomization (DR) and real-to-sim calibration. To learn locomotion behaviors robust to terrain variations, we randomize terrain roughness, friction, and restitution. To account for inaccuracies in the robot's URDF, we randomize the mass and center of mass position of each of the robot's links. We also randomize the PD gains and stall torques for each actuator in the robot's legs, and the policy lag length to learn robustness to latencies observed on hardware. To encourage learning recovery behaviors, we randomize the initial joint and body states of the robot and periodically perturb it with external forces and torques at the base, hips, feet, and end-effector, following the approach proposed in \cite{disneyBDX}. The DR ranges used for both pre-training and fine-tuning are provided in Appendix \ref{sec:appendix_training_details}. 

Inspired by \cite{shin2023actuatorconstrained}, we clip the commanded motor torques $\boldsymbol\tau$ such that 
\begin{align}
    \boldsymbol{\tau} &\geq -\boldsymbol{\tau}_{\max} \left( 1 + \max \left( \min \left( \frac{\dot{\mathbf{q}}}{\dot{\mathbf{q}}_{\max}}, 0 \right), -1 \right) \right), \\
    \boldsymbol{\tau} &\leq \boldsymbol{\tau}_{\max} \left( 1 - \max\left( \min \left( \frac{\dot{\mathbf{q}}}{\dot{\mathbf{q}}_{\max}}, 1 \right), 0 \right) \right).
\end{align}
where $\boldsymbol{\tau}_{\max}$ and $\dot{\mathbf{q}}_{\max}$ are the maximum torques and velocities of the actuators, respectively. This clipping strategy enforces a physical motor constraint by ensuring that torque commands do not demand power beyond the motor's maximum output capacity. Furthermore, we clip the arm torques a second time to satisfy the constraint
\begin{equation}
    \lvert\boldsymbol{\tau}\rvert^\top \lvert\dot{\qq} \rvert \leq P_{\rm max},
\end{equation}
where $P_{\rm max}$ is the maximum total power of the arm joints,
because we found experimentally this helps prevent the arm from entering a power protect state enforced by the robot's manufacturer.

Since typical DR strategies were insufficient for athletic behaviors in the arm (which uses harmonic drives), we incorporate the UAN (Section~\ref{sec:residual_actuator_net}) for the arm actuators. Therefore, no DR is applied to arm joint properties.

\subsubsection{Task Specification} The pre-training task for the WBC is to track a desired base velocity and EE pose. The velocity command, $\mathbf{v}_t^{\rm cmd}=\left[v^{\rm cmd}_{x,t}, v^{\rm cmd}_{y,t}, \omega^{\rm cmd}_{z,t}\right]$, consists of a desired forward velocity $v^{\rm cmd}_{x,t}$, a desired lateral velocity $v^{\rm cmd}_{y,t}$, and a desired yaw-rate $\omega^{\rm cmd}_{z,t}$. 
We command the EE pose in a yaw-rotated frame aligned with the robot's center of mass at a fixed height above the terrain. The choice of frame encourages the robot to coordinate with its legs to expand its workspace.
The EE command $\mathbf{p}_t^{\rm cmd}=\left[ p^{\rm cmd}_{{\rm EE},t}, o^{\rm cmd}_{{\rm EE},t} \right]$ comprises a cartesian position $p^{\rm EE}_t$ and orientation $o^{\rm EE}_t$ (provided as the first two columns of a rotation matrix).

\subsubsection{Reward Function} 
The reward function is split as $r_t = r^{\rm track}_t + r^{\rm aux}_t$, where $r^{\rm track}_t$ include tracking terms (EE pose, base velocity) and gait terms, while $r^{\rm aux}_t$ includes regularization and smoothing terms.
The EE tracking term rewards minimizing the distance between four key points, where one key point is positioned at the frame's origin, and the others are positioned along each axis of the frame. Full details are provided in Appendix \ref{sec:appendix_training_details}. 

\subsubsection{Command Sampling Scheme} 
We adopt the approach first proposed in \cite{fu2023deep} to sample commands during training. We sample a new base velocity command and a new goal end effector pose every 7 seconds of simulation time. Upon sampling, the command is linearly interpolated (over 2 to 5 seconds) from the previous command.
While this sampling scheme suffices for foundational loco-manipulation skills, it may be too smooth for highly agile motions -- this motivates our task-specific fine-tuning (Section \ref{sec:fine-tune}).

\subsection{Task-Specific Finetuning} \label{sec:fine-tune}
After pre-training, the policy can track reference trajectories, but struggles on high-acceleration tasks.  
To address this, we fine-tune the policy directly with task rewards.
The same WBC base policy weights can be reused for multiple task policies, thus avoiding repeated pre-training.

\subsubsection{Initialization} 
The policy weights are initialized to those learned during pre-training.
To avoid policy collapse, we set a low initial learning rate ($1 \times 10^{-5}$) for the actor and retain the standard deviation from pre-training. Additionally, we set the entropy coefficient in PPO to zero during fine-tuning to improve training stability.

\subsubsection{Reference trajectory and task embedding} During fine-tuning, the policy receives a task-specific reference trajectory and a one-hot task embedding to inform which phase of the task (e.g., set-up, execute, settle) is active.
We hand-designed the reference trajectories through joint interpolation and forward kinematics, but they could also come from an expert policy or human demonstration. 

\subsubsection{Fine-tuning with task reward} The environment for fine-tuning phase extends that of pre-training (same DR ranges, external pushes, etc.). 
The reward becomes $r^i_t + r^{\rm aux}_t$, where $r^i_t$ is task-specific.
Initially, the policy tracks the reference, aiding exploration; later, it learns to deviate to maximize task performance.

\begin{figure}
    \centering
    \includegraphics[width=0.8\linewidth]{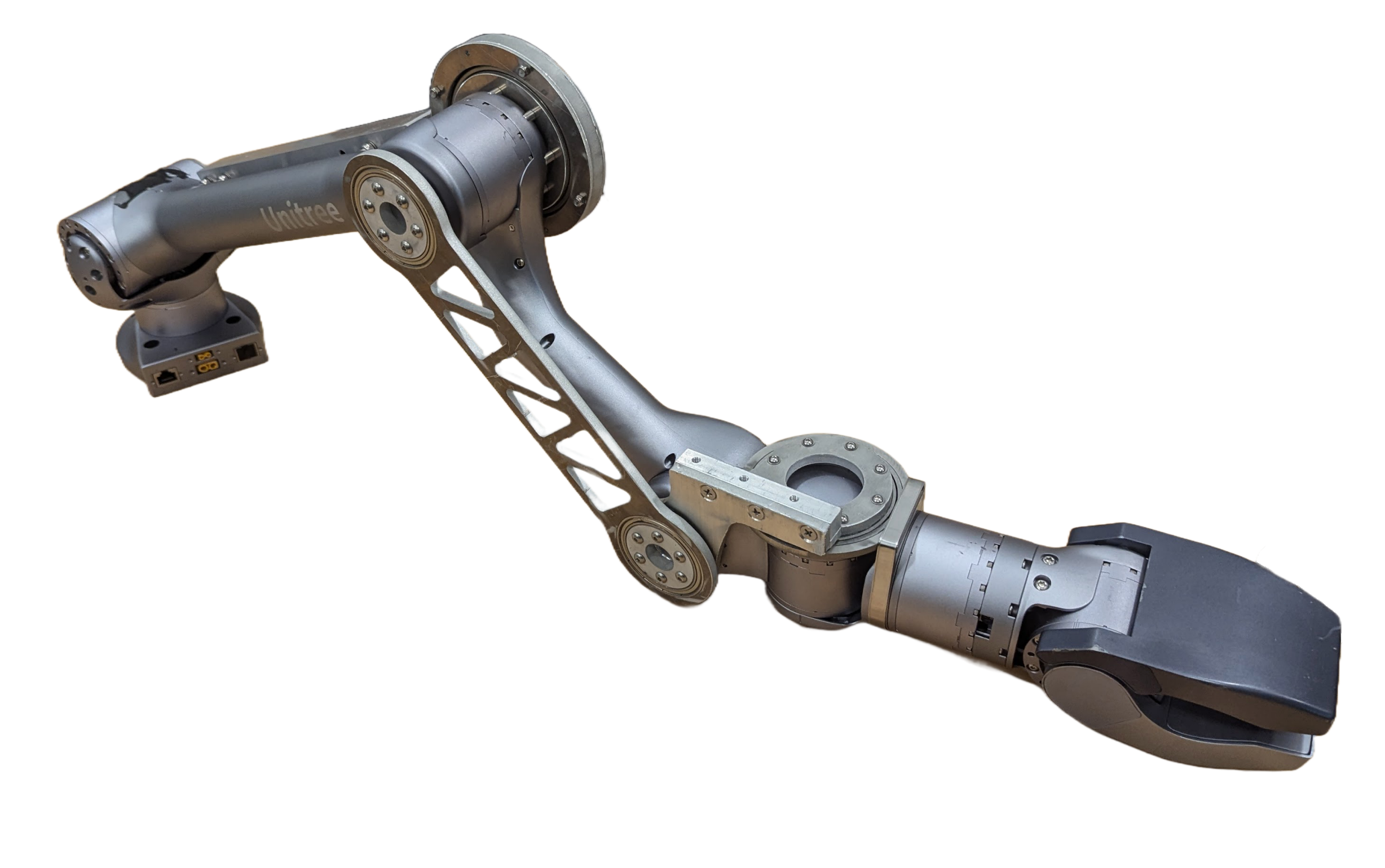}
    \caption{\textbf{Unitree Z1 Pro arm.} This arm's harmonic actuators behave substantially differently from the quasi-direct-drive motors common in small legged robots. This image also shows the reinforcements we designed to ensure that the limit on athleticism comes from actuation rather than the linkage structural integrity.}
    \label{fig:updated-arm}
\end{figure}

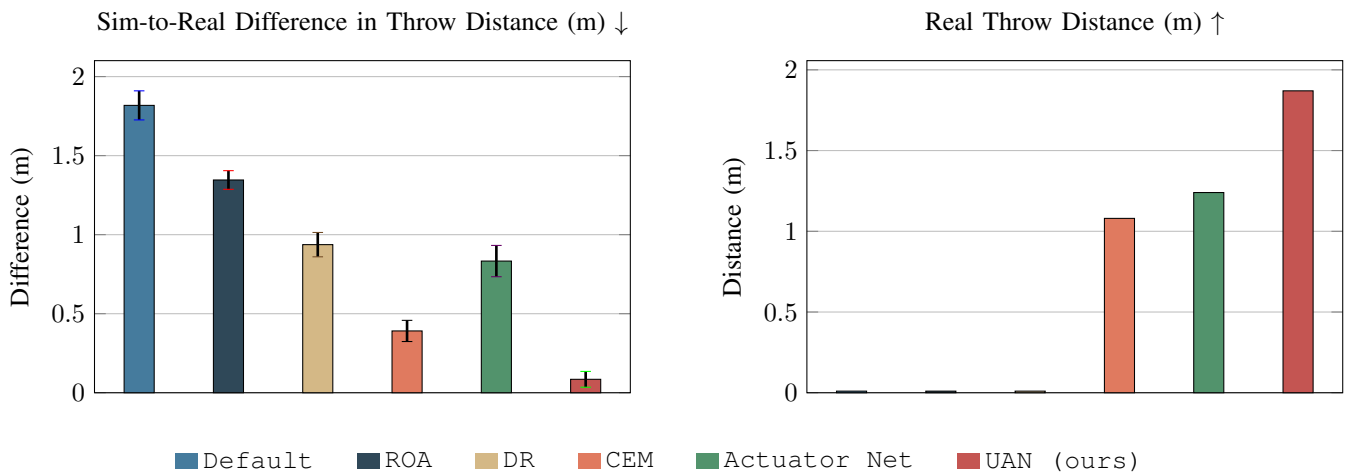
\begin{figure*}[htbp]
  \centering
  \begin{subfigure}[b]{0.48\textwidth}
    \centering
    \begin{tikzpicture}
      \begin{axis}[
         ybar,
         bar width=0.4cm,
         bar shift=0.0cm,
         width=\textwidth,
         height=6cm,
         symbolic x coords={Ideal,ROA,DR,CEM,{Actuator Net},{UAN (ours)}},
         xtick=\empty,
         xticklabel=\empty,         
         title={Sim-to-Real Difference in Throw Distance (m) $\downarrow$},
         ylabel={Difference (m)},
         ymin=0,
         clip=false,
         grid=major,
      ]
         \addplot+[ybar, fill=IdealColor, draw = black, 
           error bars/.cd, y dir=both, y explicit, 
           error bar style={draw=black, line width=1pt}]
           coordinates {(Ideal,1.818) +- (0,0.092)};
         \addplot+[ybar, fill=ROAColor, draw = black, 
           error bars/.cd, y dir=both, y explicit, 
           error bar style={draw=black, line width=1pt}]
           coordinates {(ROA,1.346) +- (0,0.059)};
         \addplot+[ybar, fill=DRColor, draw = black, 
           error bars/.cd, y dir=both, y explicit, 
           error bar style={draw=black, line width=1pt}]
           coordinates {(DR,0.937) +- (0,0.077)};
         \addplot+[ybar, fill=CEMColor, draw = black, 
           error bars/.cd, y dir=both, y explicit, 
           error bar style={draw=black, line width=1pt}]
           coordinates {(CEM,0.391) +- (0,0.067)};
         \addplot+[ybar, fill=ActuatorNetColor, draw = black,
           error bars/.cd, y dir=both, y explicit, 
           error bar style={draw=black, line width=1pt}]
           coordinates {({Actuator Net},0.833) +- (0,0.099)};
         \addplot+[ybar, fill=UANColor, draw = black, 
           error bars/.cd, y dir=both, y explicit, 
           error bar style={draw=black, line width=1pt}]
           coordinates {({UAN (ours)},0.085) +- (0,0.050)};
      \end{axis}
    \end{tikzpicture}
    
  \end{subfigure}
  \hfill
  \begin{subfigure}[b]{0.48\textwidth}
    \centering
    \begin{tikzpicture}
      \begin{axis}[
         ybar,
         bar width=0.4cm,
         bar shift=0.0cm,
         width=\textwidth,
         height=6cm,
         symbolic x coords={Ideal,ROA,DR,CEM,{Actuator Net},{UAN (ours)}},
         xtick=\empty,
         xticklabel=\empty,         
         title={Real Throw Distance (m) $\uparrow$},
         ylabel={Distance (m)},
         ymin=0,
         clip=false,
         grid=major,
      ]
         \addplot+[ybar, fill=IdealColor, draw=black]    coordinates {(Ideal,0.01)};
         \addplot+[ybar, fill=ROAColor, draw=black]       coordinates {(ROA,0.01)};
         \addplot+[ybar, fill=DRColor, draw=black]        coordinates {(DR,0.01)};
         \addplot+[ybar, fill=CEMColor, draw=black]       coordinates {(CEM,1.08)};
         \addplot+[ybar, fill=ActuatorNetColor, draw=black] coordinates {({Actuator Net},1.24)};
         \addplot+[ybar, fill=UANColor, draw=black]       coordinates {({UAN (ours)},1.87)};
      \end{axis}
    \end{tikzpicture}
  \end{subfigure}

  \vspace{1ex}
  \begin{tikzpicture}
    \begin{axis}[
       hide axis,
       scale only axis,
       height=0pt,
       width=0pt,
       axis lines=none,
       ticks=none,
       legend columns=6,
       legend style={
         draw=none,
         inner sep=0pt,
         every legend image/.append style={mark=none, mark size=0pt},
         /tikz/every even column/.append style={column sep=0.5cm},
         legend image code/.code={
           \draw[fill=##1, draw=none] (0cm,-0.1cm) rectangle (0.3cm,0.1cm);
         },
       },
    ]
      \addplot+[ybar, fill=IdealColor]    coordinates {(0,0)};
      \addlegendentry{\texttt{Default}}
      \addplot+[ybar, fill=ROAColor]       coordinates {(0,0)};
      \addlegendentry{\texttt{ROA}}
      \addplot+[ybar, fill=DRColor]        coordinates {(0,0)};
      \addlegendentry{\texttt{DR}}
      \addplot+[ybar, fill=CEMColor]       coordinates {(0,0)};
      \addlegendentry{\texttt{CEM}}
      \addplot+[ybar, fill=ActuatorNetColor] coordinates {(0,0)};
      \addlegendentry{\texttt{Actuator Net}}
      \addplot+[ybar, fill=UANColor]       coordinates {(0,0)};
      \addlegendentry{\texttt{UAN (ours)}}
    \end{axis}
    \node[anchor=north east, fill=white, draw=white, circle, minimum size=12pt, xshift=5pt, yshift=5pt] at (current bounding box.north east) {};
  \end{tikzpicture}
  
  \caption{\textbf{UAN improves simulator accuracy and real throwing performance.} 
  \texttt{UAN (Ours)} achieves lower sim-to-real difference in throw distance as compared to standard baselines, resulting in a better real throw distance. For this comparison, we train and test policies with a fixed-base arm, to avoid the risk of the legged base falling during performance-critical ablations.}
  \label{fig:throwing_metrics}
\end{figure*}

\section{Experimental Setup}

We chose the Unitree B2 with Unitree Z1 Pro arm as our hardware platform, and we consider three athletic tasks: throwing, weight lifting, and sled pulling (see Section \ref{sec:hardware_results}).  Structural upgrades to the arm were custom designed and fabricated to withstand the high loads during athletic behaviors (see Section \ref{sec:arm_modifications}).

Following our UAN training (Section \ref{sec:residual_actuator_net}), we pre-trained a WBC (Section \ref{sec:pre-train}) and then fine-tuned policies for each task (Section \ref{sec:fine-tune}). Ablations comparing our method with alternatives are described in Section \ref{sec:results_sysid_ablation} and Section \ref{sec:results_fine-tune-ablation}.

\subsection{Arm Modifications} 
\label{sec:arm_modifications}
During development, the Unitree Z1 Pro arm experienced structural failures at links 2 and 4, with minor deformations at link 5. The damage resulted from the highly dynamic movements in the athletic experiments, which applied loads to the links that exerted excessive stress and strain on the links exceeding the material's yield strength. Modifications were made to reinforce
links 2, 3, 4, and 5 by adding supports at the joints. This prevents loads from being transferred solely through the motors which are cantilevered. 
A mass-efficient aluminum square tube was used for link 2, which experiences the highest stress of all the links. Idler bearings are used to apply support at the motor outputs without restricting their movement. In the URDF, link masses, centers of mass, and inertias were updated based on CAD calculations and the parallel axis theorem. Figure \ref{fig:updated-arm} shows the reinforced arm.

\section{Experimental Results}
\label{sec:results}

In this section, we report ablations that identify the contribution of key system components and present results for the athletic tasks. Supplemental videos are provided on the project website: \url{https://uan.csail.mit.edu/}.

Our experiments address the following questions:
\begin{enumerate}
    \item Does our \textbf{unsupervised actuator net} reduce the sim-to-real gap and improve transfer?
    \item What are the benefits of our two-stage \textbf{pre-training and fine-tuning} pipeline relative to alternatives? 
    \item Does our approach enable sim-to-real transfer of \textbf{athletic whole-body control tasks}?
\end{enumerate}

\subsection{Comparing System Identification Approaches} \label{sec:results_sysid_ablation}

\begin{figure*}[htbp]
  \centering
  \begin{subfigure}[b]{0.32\textwidth}
    \centering
    \begin{tikzpicture}
      \begin{axis}[
        ybar,
         bar width=0.4cm,
         bar shift=0.0cm,
        width=\textwidth,
        height=6cm,
        ymin=0,
        symbolic x coords={No-Fine-Tuning,No-Pre-Training,No-E2E,Ours},
        xtick=\empty,
        xticklabel=\empty,         
        title={Distance Thrown (m) $\uparrow$},
        enlarge x limits=0.2,       
        clip=false,                 
        grid=major,
      ]
        \addplot+[ybar, fill=Sage, draw=black, 
          error bars/.cd, y dir=both, y explicit,
          error bar style={draw=black, fill=black, line width=1pt}]
          coordinates {(No-Fine-Tuning,0.6) +- (0,0.2)};
        \addplot+[ybar, fill=DustyRose, draw=black, 
          error bars/.cd, y dir=both, y explicit,
          error bar style={draw=black, fill=black, line width=1pt}]
          coordinates {(No-Pre-Training,10.2) +- (0,0.7)};
        \addplot+[ybar, fill=SlateBlue, draw=black, 
          error bars/.cd, y dir=both, y explicit,
          error bar style={draw=black, fill=black, line width=1pt}]
          coordinates {(No-E2E,10.9) +- (0,0.6)};
        \addplot+[ybar, fill=WarmTaupe, draw=black, 
          error bars/.cd, y dir=both, y explicit,
          error bar style={draw=black, fill=black, line width=1pt}]
          coordinates {(Ours,15.1) +- (0,1.5)};
      \end{axis}
    \end{tikzpicture}
  \end{subfigure}
  \hfill
  \begin{subfigure}[b]{0.32\textwidth}
    \centering
    \begin{tikzpicture}
      \begin{axis}[
        ybar,
         bar width=0.4cm,
         bar shift=0.0cm,
        width=\textwidth,
        height=6cm,
        ymin=0,
        symbolic x coords={No-Fine-Tuning,No-Pre-Training,No-E2E,Ours},
        xtick=\empty,
        xticklabel=\empty,
        title={Throw Release Speed (m/s) $\uparrow$},
        enlarge x limits=0.2,
        clip=false,
        grid=major,
      ]
        \addplot+[ybar, fill=Sage, draw=black, 
          error bars/.cd, y dir=both, y explicit,
          error bar style={draw=black, fill=black, line width=1pt}]
          coordinates {(No-Fine-Tuning,1.8) +- (0,0.3)};
        \addplot+[ybar, fill=DustyRose, draw=black, 
          error bars/.cd, y dir=both, y explicit,
          error bar style={draw=black, fill=black, line width=1pt}]
          coordinates {(No-Pre-Training,9.5) +- (0,0.4)};
        \addplot+[ybar, fill=SlateBlue, draw=black, 
          error bars/.cd, y dir=both, y explicit,
          error bar style={draw=black, fill=black, line width=1pt}]
          coordinates {(No-E2E,9.6) +- (0,0.3)};
        \addplot+[ybar, fill=WarmTaupe, draw=black, 
          error bars/.cd, y dir=both, y explicit,
          error bar style={draw=black, fill=black, line width=1pt}]
          coordinates {(Ours,11.7) +- (0,0.5)};
      \end{axis}
    \end{tikzpicture}
  \end{subfigure}
  \hfill
  \begin{subfigure}[b]{0.32\textwidth}
    \centering
    \begin{tikzpicture}
      \begin{axis}[
        ybar,
         bar width=0.4cm,
         bar shift=0.0cm,
        width=\textwidth,
        height=6cm,
        ymin=0,
        symbolic x coords={No-Fine-Tuning,No-Pre-Training,No-E2E,Ours},
        xtick=\empty,
        xticklabel=\empty,
        title={Peak Leg Power (kW) $\downarrow$},
        enlarge x limits=0.2,
        clip=false,
        grid=major,
      ]
        \addplot+[ybar, fill=Sage, draw=black, 
          error bars/.cd, y dir=both, y explicit,
          error bar style={draw=black, fill=black, line width=1pt}]
          coordinates {(No-Fine-Tuning,9.278) +- (0,1.391)};
        \addplot+[ybar, fill=DustyRose, draw=black, 
          error bars/.cd, y dir=both, y explicit,
          error bar style={draw=black, fill=black, line width=1pt}]
          coordinates {(No-Pre-Training,16.022) +- (0,0.252)};
        \addplot+[ybar, fill=SlateBlue, draw=black, 
          error bars/.cd, y dir=both, y explicit,
          error bar style={draw=black, fill=black, line width=1pt}]
          coordinates {(No-E2E,15.043) +- (0,0.866)};
        \addplot+[ybar, fill=WarmTaupe, draw=black, 
          error bars/.cd, y dir=both, y explicit,
          error bar style={draw=black, fill=black, line width=1pt}]
          coordinates {(Ours,13.714) +- (0,0.794)};
      \end{axis}
    \end{tikzpicture}
  \end{subfigure}
  
  \vspace{1ex}
  \begin{tikzpicture}
    \begin{axis}[
      hide axis,
      scale only axis,
      height=0pt,
      width=0pt,
      legend columns=4,
      legend style={
        draw=none,
        /tikz/every even column/.append style={column sep=1cm},
        legend image code/.code={
           \draw[fill=##1, draw=none] (0cm,-0.1cm) rectangle (0.3cm,0.1cm);
         },
      },
    ]
      \addplot+[ybar, fill=Sage] coordinates {(0,0)};
      \addlegendentry{\texttt{No-Fine-Tuning}}
      \addplot+[ybar, fill=DustyRose] coordinates {(0,0)};
      \addlegendentry{\texttt{No-Pre-Training}}
      \addplot+[ybar, fill=SlateBlue] coordinates {(0,0)};
      \addlegendentry{\texttt{No-E2E}}
      \addplot+[ybar, fill=WarmTaupe] coordinates {(0,0)};
      \addlegendentry{\texttt{Ours}}
    \end{axis}
    \node[anchor=north east, fill=white, draw=white, circle, minimum size=12pt, xshift=5pt, yshift=5pt] at (current bounding box.north east) {};
  \end{tikzpicture}
  
  \caption{\textbf{End-to-end fine-tuning from a pre-trained WBC leads to the best task performance.} Throwing evaluation metrics across $100$ simulated throws for four policies: Our fine-tuned WBC (\texttt{Ours}) achieves the longest throw distance with lower peak leg power as compared to a throwing policy trained from scratch (\texttt{No-Pre-Training}) or a high-level policy for a frozen WBC (\texttt{No-E2E}). The WBC before finetuning (\texttt{No-Fine-Tuning}) has the lowest peak leg power but throws the ball a much shorter distance.}
  \label{fig:fine-tune-ablation}
\end{figure*}
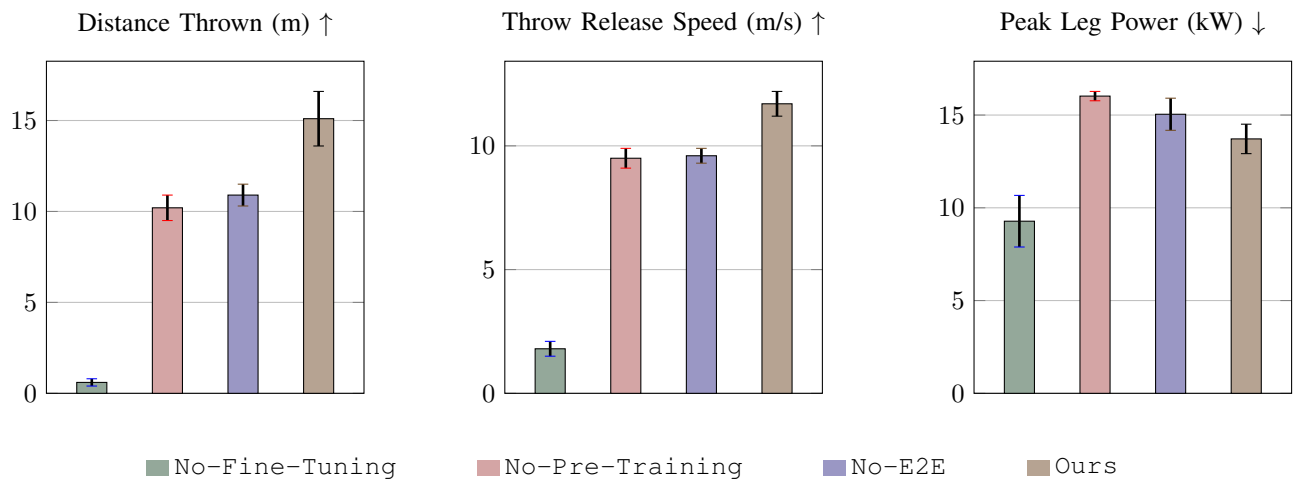

We compare several methods for modeling the actuator dynamics of the Unitree Z1 Pro arm in Isaac Sim. In particular, we consider:

\begin{enumerate}
    \item \textbf{\texttt{Default}}: The baseline simulator with no additional modifications. 
    \item \textbf{\texttt{DR}}: The simulator augmented with domain randomization (randomizing PD gains, friction, and armature parameters).
    \item \textbf{\texttt{ROA}}: A domain randomization baseline enhanced with an online system identification module via Regularized Online Adaptation~\cite{fu2023deep}.
    \item \textbf{\texttt{Actuator Net}}: A supervised actuator network following Hwangbo et al.~\cite{Hwangbo_2019} where torque labels are estimated from the motor current. (Note that these labels do not capture the nonlinear effects introduced by the harmonic reducers.)
    \item \textbf{\texttt{CEM}}: A method in which friction, frictional damping, and armature parameters are optimized using the cross-entropy method to minimize the mean-square joint position error between simulation and hardware.
    \item \textbf{\texttt{UAN}}: Our proposed unsupervised actuator network that learns corrective torques without requiring torque supervision, thereby capturing both lag and nonlinearities from harmonic reduction.
\end{enumerate}

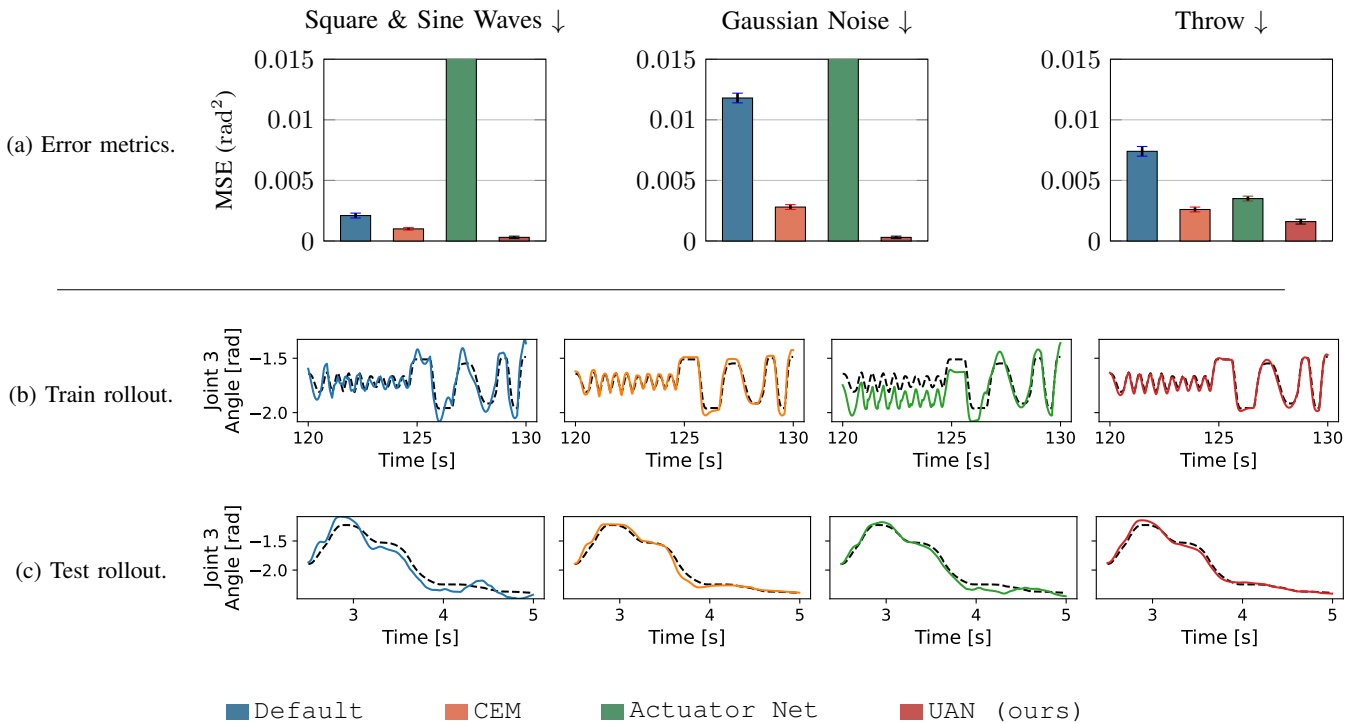
\begin{figure*}[htbp]
  \centering
   \centering
  \begin{subfigure}[b]{0.15\textwidth}
  \begin{minipage}[c]{\textwidth}
            \vspace{-3cm}  
            \caption{Error metrics.}
            \label{fig:sysid_comparison_quantitative}
        \end{minipage}%
  \end{subfigure}
  \begin{subfigure}[b]{0.25\textwidth}
    \centering
    \begin{tikzpicture}
      \begin{axis}[
        ybar,
         bar width=0.4cm,
         bar shift=0.0cm,
        width=\textwidth,
        height=4cm,
        ymin=0,
        ymax=0.015,
        symbolic x coords={Default,CEM,Actuator Net,UAN (ours)},
        xtick=\empty,
        xticklabel=\empty,         
        title={Square \& Sine Waves $\downarrow$},
        ylabel={MSE ($\SI{}{\radian ^2}$)},
        ylabel style={font=\small},  
        enlarge x limits=0.2,       
        clip=true,                 
        grid=major,
        scaled y ticks = false,
        yticklabel style={/pgf/number format/fixed,/pgf/number format/precision=3},
      ]
        \addplot+[ybar, fill=IdealColor, draw=black, 
          error bars/.cd, y dir=both, y explicit,
          error bar style={draw=black, fill=black, line width=1pt}]
          coordinates {(Default,0.0021) +- (0,0.0002)};
        \addplot+[ybar, fill=CEMColor, draw=black, 
          error bars/.cd, y dir=both, y explicit,
          error bar style={draw=black, fill=black, line width=1pt}]
          coordinates {(CEM,0.0010) +- (0,0.0001)};
        \addplot+[ybar, fill=ActuatorNetColor, draw=black, 
          error bars/.cd, y dir=both, y explicit,
          error bar style={draw=black, fill=black, line width=1pt}]
          coordinates {(Actuator Net,0.0180) +- (0,0.0006)};
        \addplot+[ybar, fill=UANColor, draw=black, 
          error bars/.cd, y dir=both, y explicit,
          error bar style={draw=black, fill=black, line width=1pt}]
          coordinates {(UAN (ours),0.0003) +- (0,0.0001)};
      \end{axis}
    \end{tikzpicture}
  \end{subfigure}
  \hfill
  \begin{subfigure}[b]{0.25\textwidth}
    \centering
    \begin{tikzpicture}
      \begin{axis}[
        ybar,
         bar width=0.4cm,
         bar shift=0.0cm,
        width=\textwidth,
        height=4cm,
        ymin=0,
        ymax=0.015,
        symbolic x coords={Default,CEM,Actuator Net,UAN (ours)},
        xtick=\empty,
        xticklabel=\empty,
        title={Gaussian Noise $\downarrow$},
        enlarge x limits=0.2,
        clip=true,
        grid=major,
        scaled y ticks = false,
        yticklabel style={/pgf/number format/fixed,/pgf/number format/precision=3},
      ]
        \addplot+[ybar, fill=IdealColor, draw=black, 
          error bars/.cd, y dir=both, y explicit,
          error bar style={draw=black, fill=black, line width=1pt}]
          coordinates {(Default,0.0118) +- (0,0.0004)};
        \addplot+[ybar, fill=CEMColor, draw=black, 
          error bars/.cd, y dir=both, y explicit,
          error bar style={draw=black, fill=black, line width=1pt}]
          coordinates {(CEM,0.0028) +- (0,0.0002)};
        \addplot+[ybar, fill=ActuatorNetColor, draw=black, 
          error bars/.cd, y dir=both, y explicit,
          error bar style={draw=black, fill=black, line width=1pt}]
          coordinates {(Actuator Net,0.1252) +- (0,0.0013)};
        \addplot+[ybar, fill=UANColor, draw=black, 
          error bars/.cd, y dir=both, y explicit,
          error bar style={draw=black, fill=black, line width=1pt}]
          coordinates {(UAN (ours),0.0003) +- (0,0.0001)};
      \end{axis}
    \end{tikzpicture}
  \end{subfigure}
  \hfill
  \begin{subfigure}[b]{0.25\textwidth}
    \centering
    \begin{tikzpicture}
      \begin{axis}[
        ybar,
         bar width=0.4cm,
         bar shift=0.0cm,
        width=\textwidth,
        height=4cm,
        ymin=0,
        ymax=0.015,
        symbolic x coords={Default,CEM,Actuator Net,UAN (ours)},
        xtick=\empty,
        xticklabel=\empty,
        title={Throw $\downarrow$},
        enlarge x limits=0.2,
        clip=true,
        grid=major,
        scaled y ticks = false,
        yticklabel style={/pgf/number format/fixed,/pgf/number format/precision=3},
      ]
        \addplot+[ybar, fill=IdealColor, draw=black, 
          error bars/.cd, y dir=both, y explicit,
          error bar style={draw=black, fill=black, line width=1pt}]
          coordinates {(Default,0.0074) +- (0,0.0004)};
        \addplot+[ybar, fill=CEMColor, draw=black, 
          error bars/.cd, y dir=both, y explicit,
          error bar style={draw=black, fill=black, line width=1pt}]
          coordinates {(CEM,0.0026) +- (0,0.0002)};
        \addplot+[ybar, fill=ActuatorNetColor, draw=black, 
          error bars/.cd, y dir=both, y explicit,
          error bar style={draw=black, fill=black, line width=1pt}]
          coordinates {(Actuator Net,0.0035) +- (0,0.0002)};
        \addplot+[ybar, fill=UANColor, draw=black, 
          error bars/.cd, y dir=both, y explicit,
          error bar style={draw=black, fill=black, line width=1pt}]
          coordinates {(UAN (ours),0.0016) +- (0,0.0002)};
      \end{axis}
    \end{tikzpicture}
  \end{subfigure}

    
    \rule{0.9\linewidth}{0.2pt}
    
    \vspace{1.5em}
   
    \begin{subfigure}[b]{\textwidth}
        \begin{minipage}[c]{0.15\textwidth}
            \vspace{-\baselineskip}  
            \caption{Train rollout.}
            \label{fig:sysid_comparison_j3_train}
        \end{minipage}%
        \begin{minipage}[c]{0.85\textwidth}
            \includegraphics[width=\textwidth,trim=0cm 0cm 0cm 0cm,clip]{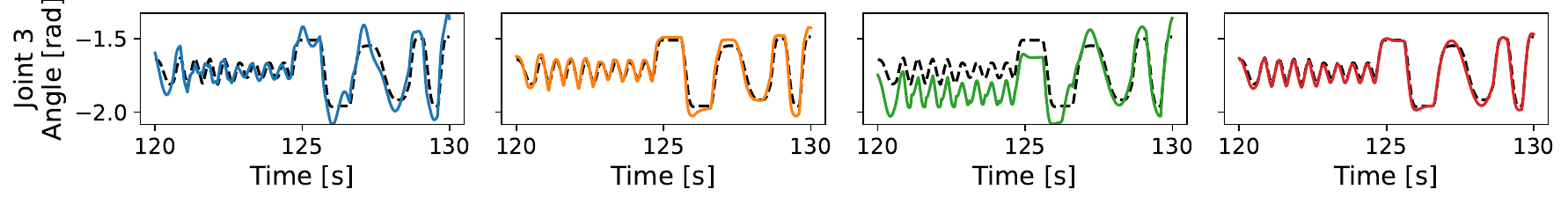}
        \end{minipage}
    \end{subfigure}
    
    \vspace{1em} 

    \begin{subfigure}[b]{\textwidth}
        \begin{minipage}[c]{0.15\textwidth}
            \vspace{-\baselineskip}  
            \caption{Test rollout.}
            \label{fig:sysid_comparison_j3_test}
        \end{minipage}%
        \begin{minipage}[c]{0.85\textwidth}
            \includegraphics[width=\textwidth,trim=0cm 1.2cm 0cm 0cm,clip]{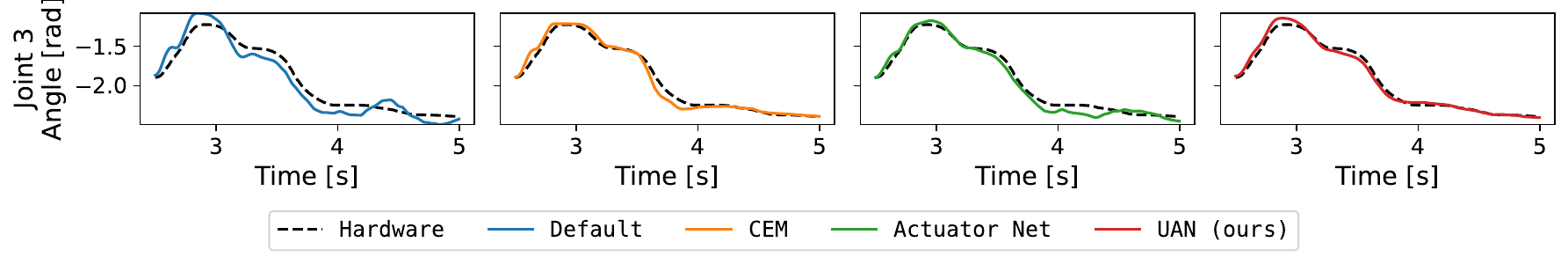}
        \end{minipage}
    \end{subfigure}

  \vspace{1ex}
  \begin{tikzpicture}
    \begin{axis}[
      hide axis,
      scale only axis,
      height=0pt,
      width=0pt,
      legend columns=4,
      legend style={
        draw=none,
        /tikz/every even column/.append style={column sep=1cm},
        legend image code/.code={
           \draw[fill=##1, draw=none] (0cm,-0.1cm) rectangle (0.3cm,0.1cm);
         },
      },
    ]
      \addplot+[ybar, fill=IdealColor] coordinates {(0,0)};
      \addlegendentry{\texttt{Default}}
      \addplot+[ybar, fill=CEMColor] coordinates {(0,0)};
      \addlegendentry{\texttt{CEM}}
      \addplot+[ybar, fill=ActuatorNetColor] coordinates {(0,0)};
      \addlegendentry{\texttt{Actuator Net}}
      \addplot+[ybar, fill=UANColor] coordinates {(0,0)};
      \addlegendentry{\texttt{UAN (ours)}}
    \end{axis}
    \node[anchor=north east, fill=white, draw=white, circle, minimum size=12pt, xshift=5pt, yshift=5pt] at (current bounding box.north east) {};
  \end{tikzpicture}
  
  \caption{\textbf{UAN achieves the tightest real-to-sim fit to the training data, as well as a throw trajectory unseen during training.} We rolled out three real-world joint trajectories: square \& sine waves at each joint, Gaussian noise across all the joints, and a throw. Square waves, sine waves, and gaussian noise were seen during training, while the throw was not. We found that \texttt{Actuator-Net} error remains bounded on the \SI{5}{\second} throw trajectory but diverges when rolling out the \SI{5}{\minute} training trajectories, while the \texttt{UAN} learned to remain accurate across long rollouts through RL training.
  }
  \label{fig:sysid_comparison_metrics}
\end{figure*}

We first evaluate the modeling accuracy of these approaches by reporting the mean-square joint position error on both the training data and on an unseen test trajectory (see Figure~\ref{fig:sysid_comparison_metrics}). Our results show that the \texttt{UAN} method achieves the best fit, suggesting excellent generalization. For example, detailed windows of simulator rollouts for a single arm joint are provided in Figures~\ref{fig:sysid_comparison_j3_train} (training data) and~\ref{fig:sysid_comparison_j3_test} (test data); additional results for the other arm joints are included in the appendix (Figures~\ref{fig:sysid_comparison_train} and~\ref{fig:sysid_comparison_test}). In our observations, the \texttt{CEM} method helps prevent overshoot (by effectively slowing the arm to match the lower joint velocities seen on hardware). \texttt{Actuator Net} can improve over the baseline by capturing lag effects, but it diverged on the \SI{5}{min} rollouts on the training data. However, only \texttt{UAN} achieves a tight fit to the training data, thanks to its capacity to model the nonlinear effects introduced by the harmonic reducers. A shown by Figure~\ref{fig:sysid_comparison_metrics}, \texttt{UAN} can even accurately capture the arm's response to Gaussian noise control input, which is commonly used for exploration in reinforcement learning but represents a challenging regime for accurate simulation where the baseline methods degrade substantially.

To further assess these system identification methods in a task context, we trained arm-only throwing policies in simulation augmented with each approach and deployed them on hardware. The average throwing performance in simulation and reality is presented in Figure~\ref{fig:throwing_metrics}. In simulation, although the \texttt{Actuator Net} and \texttt{CEM} produced a promising throw, its behavior did not transfer as well to hardware. In contrast, the \texttt{UAN} policy achieved the farthest throws on hardware with the smallest sim-to-real gap. Meanwhile, the \texttt{Default}, \texttt{DR}, and \texttt{ROA} policies produced unstable behaviors—the \texttt{Default} policy, for instance, strayed excessively and failed to throw the ball at all.

\subsection{Finetuning Foundational WBC } \label{sec:results_fine-tune-ablation}
We compare four throwing policies to assess the impact of our pre-training and fine-tuning approaches:
\begin{enumerate}
    \item \textbf{\texttt{No-Fine-Tuning}}: a pre-trained WBC that tracks a throwing reference trajectory.
    \item \textbf{\texttt{No-Pre-Training}}: a throwing policy trained from scratch.
    \item \textbf{\texttt{No-E2E}}: a high-level policy 
    that outputs commands
    for a frozen pre-trained WBC.
    \item \textbf{\texttt{Ours}}: our method 
    that initializes with the pre-trained WBC and fine-tunes with RL.
\end{enumerate}
All methods observe a hand-designed throwing reference trajectory. 

Figure \ref{fig:throw-comparison} presents the performance of each throwing policy across 100 simulated throws.
\texttt{No-Fine-Tuning} successfully throws the ball by tracking the reference, but its performance is sub-optimal. However, the strong performance of \texttt{No-E2E} shows that the WBC's performance can be improved by providing a better reference trajectory. Still, the \texttt{No-E2E} policy does not perform to the maximum capability of the hardware. Through RL finetuning with the task reward, \texttt{Ours} can learn to throw farther while using a reduced peak power output in its leg motors. While \texttt{No-Pre-Training} could theoretically push the capabilities of the hardware, in practice, it struggles to do so due to exploration challenges. We found that \texttt{No-Pre-Training} achieved similar throwing performance to \texttt{No-E2E}, despite hitting a larger peak power output in its legs.  

\subsection{Hardware Results} \label{sec:hardware_results}

\subsubsection{Ball throwing} 
\begin{figure}
  \centering
    \centering
    \includegraphics[width=\linewidth]{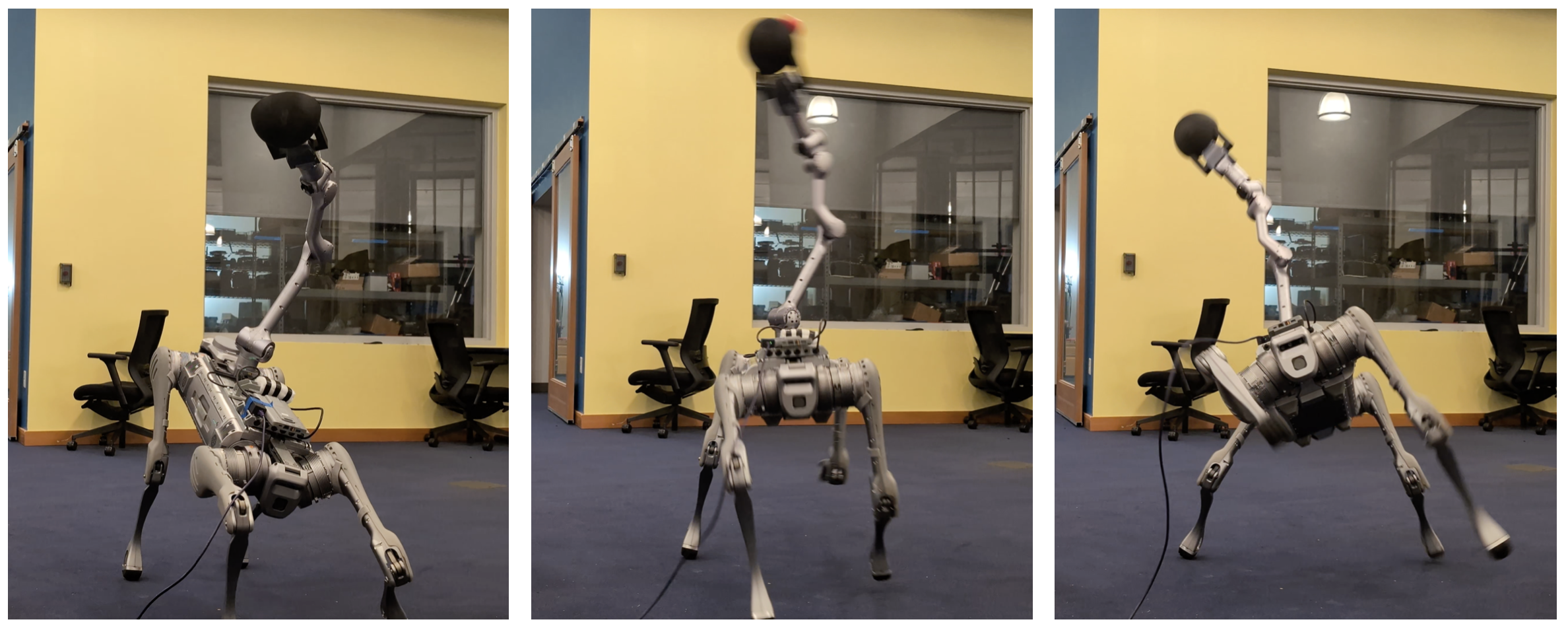}
    \includegraphics[width=\linewidth]{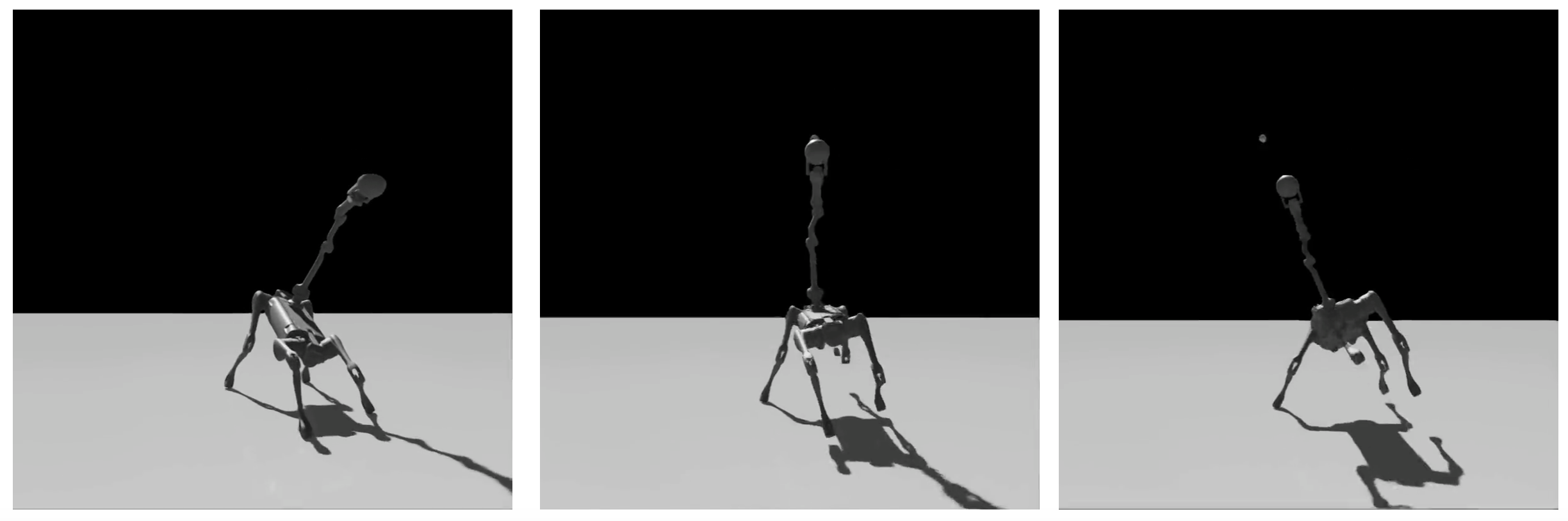}
    (a) Ball Throw 
    
    \vspace{0.2cm}
    
    \centering
    \includegraphics[width=\linewidth]{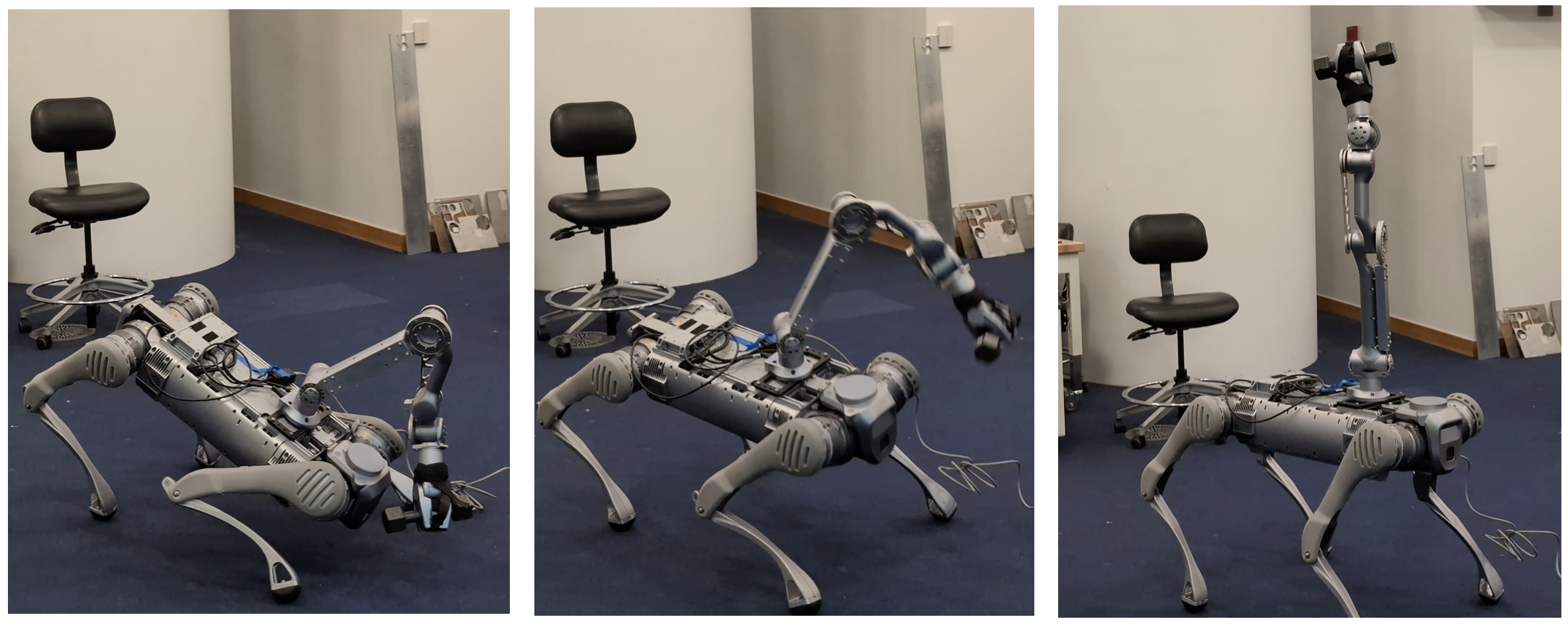}
    \includegraphics[width=\linewidth]{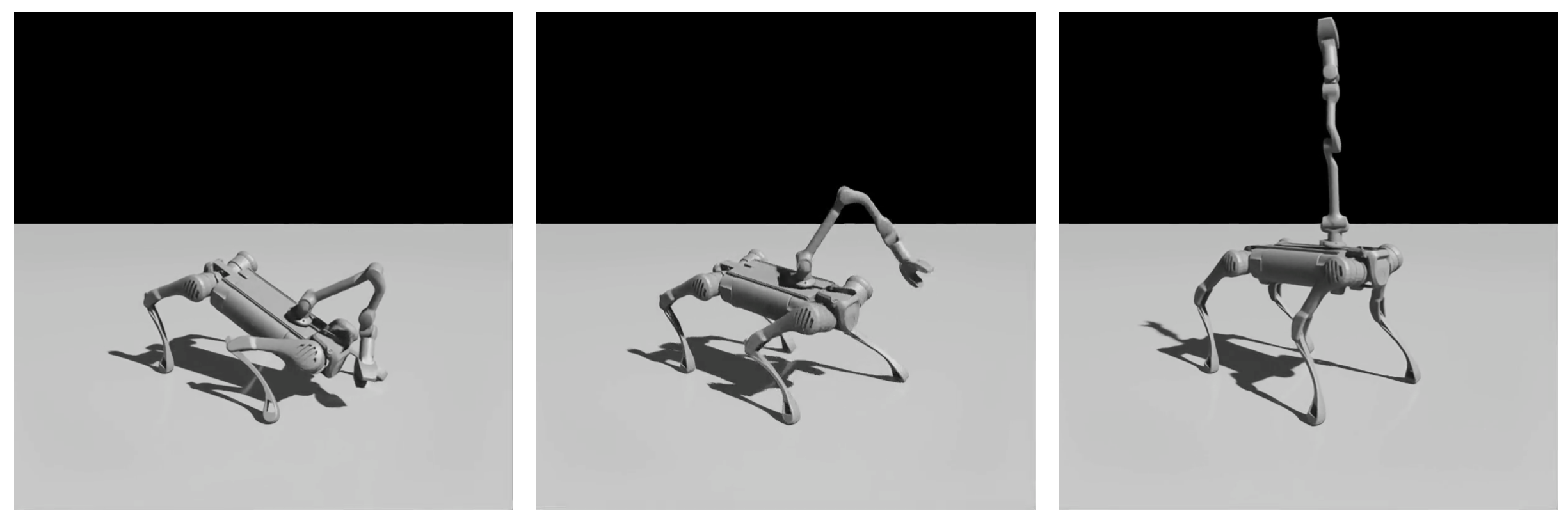}
    (b) Dumbbell Snatch 
    
    \vspace{0.2cm}
    
  
  
      \centering
      \includegraphics[width=\linewidth]{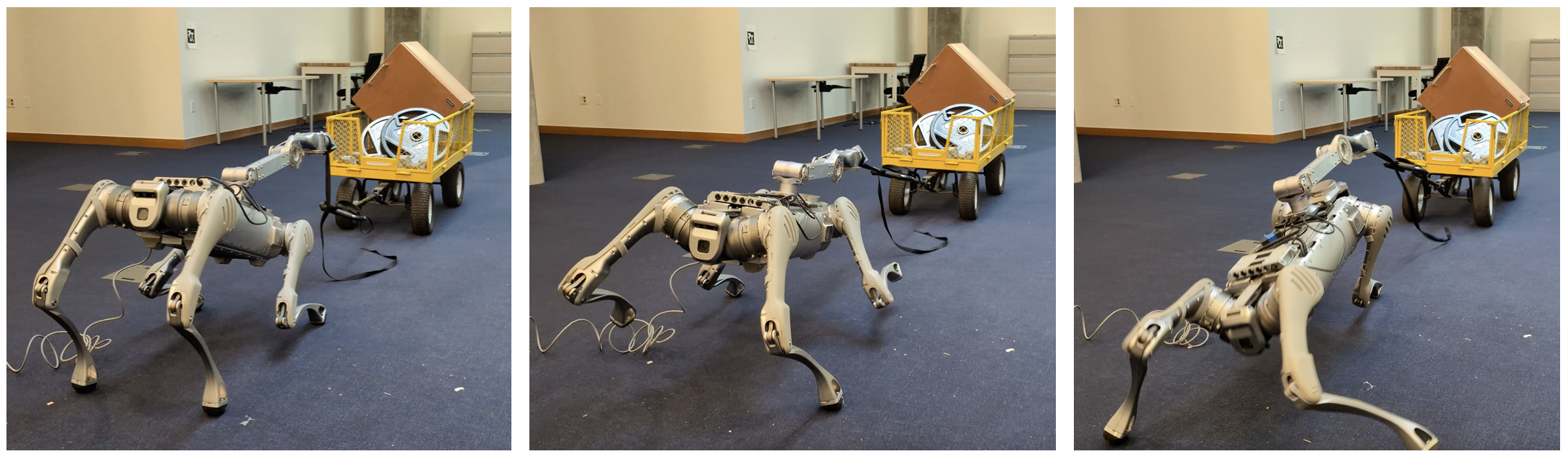}
    \includegraphics[width=\linewidth]{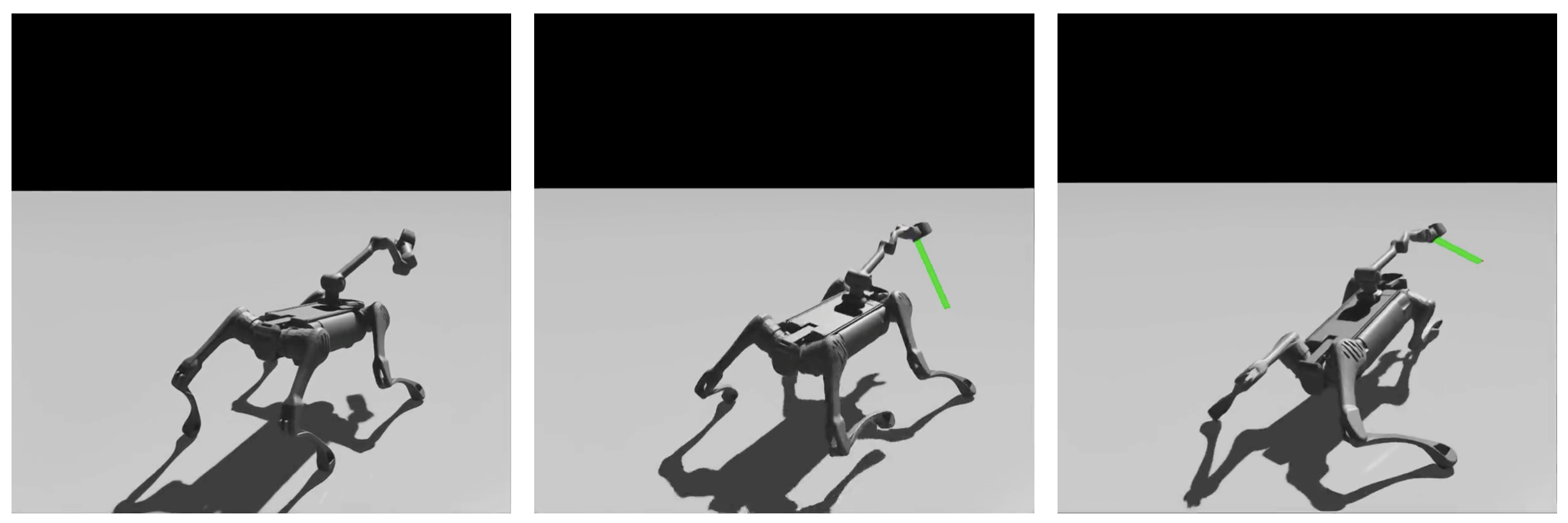}
      (c) Sled Pull
    
    \vspace{0.2cm}
    
  \caption{\textbf{Real and simulated snapshots of athletic tasks.} Visualizing simulated and real rollouts of whole-body behaviors. 
  }
  \label{fig:robot_sequences}
\end{figure}

The task objective is to throw a \SI{100}{\gram} ball as far as possible. Because grasping and releasing a ball directly is challenging for our gripper, a small bucket is attached to the robot's EE. The policy leans back prior to throwing, then pushes with its hind legs while swinging its arm forward to launch the ball. Figure \ref{fig:robot_sequences} shows side-by-side snapshots from simulation and hardware. On hardware, the ball was thrown approximately \SI{20}{\meter}, with the real robot throwing slightly further than in simulation -- possibly due to inaccuracies in the ball-bucket contact modeling.

\subsubsection{Dumbbell snatch} 
The goal is to lift a dumbbell with the EE and hold it stably.
The dumbbell is simulated by modifying the gripper's mass. 
The robot 
first lowers its EE to the ground, at which point the mass is added to its gripper. Then, its commanded to lift the weight in the air. When lifting, the robot is rewarded for maximizing the z position of its EE.

When training the lifting policy, we randomized the mass of the robot's EE from 0 to 10 kg. At convergence, the policy could consistently lift weights up to 8kg, but struggled to stabilize heavier weights above its body. Since the robot's arm is much weaker than the legs, the policy learns to pitch its base backwards to swing the weight upwards into the air. Figure \ref{fig:robot_sequences} includes snapshots of the learned lifting behavior in simulation and reality. During hardware experiments, we secured the dumbbells inside the robot's gripper with a belt to prevent it from slipping out of the robot's grasp, We found the Z1 arm could not lift even a 5 lb. dumbbell to an upright position through simple joint interpolation. We first verified the whole-body policy could lift a 5 lb. dumbbell and then progressed to a 10 lb. dumbbell. In both experiments, the robot lifted the weight above its base and maintained it there stably for over \SI{5}{\second}. 

\subsubsection{Sled pull} 
In this task, the robot pulls a heavy sled attached by a rope to its EE.
The sled is modeled as avirtual, 3-dimensional mass-spring-damper system. 
The robot is rewarded for tracking a backward base velocity while minimizing lateral drift. The policy learns to adopt a low stance to maintain balance and extend its arm to avoid applying unnecessary torques to the arm's actuators. 
In simulation, policies successfully pulled weights up to \SI{150}{\kilo\gram}. On hardware, the robot pulled a cart resisting a friction force of \SI{113}{\newton} over $10$ meters; a heavier cart (requiring \SI{230}{\newton}) was only pulled about 0.5 meters.

\section{Limitations} 
\label{sec:limitations}

Our fine-tuning approach requires a task reference trajectory, which may not be available for all robot morphologies or tasks. It also necessitates per-task engineering of the training environment (reward functions, object simulation, etc.). Future work might employ generative models to automatically synthesize task references. Additionally, our unsupervised actuator net focuses on the arm actuators. Extending real-to-sim calibration to other robot subsystems and modeling structural integrity are promising future directions.

\section{Related Work}
\label{sec:related_work}

\subsection{Whole-Body Control} 
Walking robots with arms present a formidable challenge for control due to their many degrees of freedom and complex dynamics. A typical paradigm is to implement a \textit{WBC} that optimizes actuation to achieve control objectives considering a model of the robot's kinematics and dynamics~\cite{sentis2006whole}. WBC approaches based on offline trajectory optimization or online optimization with reduced-order models have achieved considerable success in dynamic walking and manipulation~\cite{abe2013dynamic, bellicoso2019alma, murphy2012high, sleiman2021unified}. 
Recently, reinforcement learning in simulation has enabled whole-body control that can naturally handle model uncertainty, e.g. uncertain terrain and robot properties ~\cite{fu2023deep}. 
In the case of reinforcement learning-based whole-body control, the controller is a neural network that is commanded with an input reference position~\cite{cheng2024expressive, fu2023deep}, force~\cite{portela2024learning}, or whole-body pose~\cite{dugar2024learning, he2024learning, luo2023perpetual, luo2024smplolympics} and outputs joint-space actions.  

It is common to teleoperate legged-armed robots by parsing a reference trajectory from a human's movements in real-time and tracking it with a WBC; such an approach can accomplish expressive~\cite{cheng2024expressive}, forceful~\cite{portela2024learning}, or dexterous~\cite{ fu2024humanplushumanoidshadowingimitation} tasks. 
One may also train a high-level policy to select reference trajectories or a latent representation autonomously in place of the teleoperator, using either learning from demonstration \cite{ fu2024humanplushumanoidshadowingimitation, ha2024umi} or reinforcement learning \cite{liu2024visual, luo2024smplolympics}. 
However, some tasks may not be achievable by any choice of reference trajectory if they require a motion outside the training distribution of the WBC. 
It is challenging to formulate a generic pre-training scheme for whole-body control that anticipates all kinds of tasks one might want to perform  
for humanoids, motion capture datasets can provide diverse feasible reference commands~\cite{luo2023perpetual}, but for quadruped manipulators, pre-training commonly defaults 
to tracking procedurally generated smooth trajectories within the workspace~\cite{fu2023deep}. 

To avoid the reliance on high-quality pre-training, another possibility is to discard the explicit notion of reference trajectories altogether and directly train end-to-end policies for specific tasks such as fall recovery~\cite{ma2022combining}, door opening~\cite{schwarke2023curiosity}, or soccer \cite{Haarnoja_2024, Ji2022Soccer, Ji2023DribbleBot}. This enables the policy to learn highly dynamic motions to optimize the task reward, but, in practice, these motions can be hard to find due to fundamental exploration challenges in RL. We address this challenge by initializing the policy with pre-trained WBC weights and a reference trajectory.

\subsection{Overcoming the sim-to-real gap}
Prior work proposed simulated athletic tasks as a benchmark for learned whole-body control \cite{sferrazza2024humanoidbenchsimulatedhumanoidbenchmark, luo2024smplolympics}, though they left sim-to-real transfer as future work. In contrast, other studies have demonstrated sim-to-real transfer of athletic tasks on small robots with transparent actuators \cite{Haarnoja_2024, Ji2022Soccer, Ji2023DribbleBot}. Achieving sim-to-real transfer for athletic behaviors on large robots with non-ideal actuators is especially challenging because even minor modeling discrepancies can lead to reward hacking. To address this, we introduce UAN, which leverages real-world data to bridge the sim-to-real gap.

DR is a common strategy to mitigate discrepancies between simulation and reality \cite{kumar2021rmarapidmotoradaptation, Lee_2020}. In the field of dynamic legged robots, common parameters to randomize include the proportional and derivative gains of each joint, the stall torques, the link masses and inertias, and terrain properties \cite{Lee_2020, zhang2024wococo}. Excessive DR can reduce peak performance if the policy cannot identify key parameters of the environment necessary to optimize its reward function. To overcome this challenge, previous work employed teacher-student frameworks, where a student policy learns to imitate an expert policy that has access to privileged observations related to its environment \cite{fu2023deep, kumar2021rmarapidmotoradaptation, Lee_2020}. Alternatively, the policy may learn online system identification directly from an observation history. Some policy architectures (i.e., CNNs \cite{li2024reinforcementlearningversatiledynamic} and transformers \cite{radosavovic2023realworldhumanoidlocomotionreinforcement}) have been shown to achieve in-context adaptation
without relying on a teacher-student distillation.

Accurate system identification can reduce reliance on DR by mitigating the sim-to-real gap directly.
Methods for identifying inertial properties typically rely on least-squares estimation \cite{AtkesonEstimation}, including a notable approach that leverages insights about the geometric structure of the robot's dynamics to provide robustness against local optima \cite{LeeGeometric}. This method was applied to identify the inertial parameters of the MIT Humanoid \cite{SchwendemanHumanoidSysID}. In our work, we rely on the inertial properties provided in the manufacturer's URDF file.

Actuator modeling methods traditionally rely on parameterized physics models to capture effects such as static friction, dynamic friction, and reflected inertia \cite{disneyBDX}, the last of which can be set through the ``armature" setting in physics simulations such as Isaac Sim \cite{makoviychuk2021isaac} and MuJoCo \cite{mujoco}. This approach can be insufficient for actuators with complex transmission mechanisms. To address this, Hwangbo et al. \cite{Hwangbo_2019} proposed learning an actuator net, which is a neural network trained to predict an actuator's output torque from a history of position and velocity errors. The actuator net was added to the simulator during policy training to reduce the sim-to-real gap in ANYmal's series elastic actuators. Their approach, however, relies on torque sensing, which is uncommon in consumer robotic hardware. Schwendeman et al. \cite{SchwendemanHumanoidSysID} avoided reliance on an output torque sensor when training an actuator net by measuring the torques from current. However, this is only accurate in low-reduction and low-torque-density actuators which are efficiently backdriveable and have minimal reflected inertia. In contrast, our approach, UAN, employs an actuator net without relying on torque data. Instead, we train the network to predict corrective torques for the simulator that minimize the discrepancy between the simulated and real-world transition dynamics. 

When ground-truth labels are unavailable (i.e., the robot's actuators lack torque sensing), 
they can be discovered through interaction
to better match the real-world dynamics.
For example, \citet{zeng2020tossingbot} learned a residual model to better predict the ballistic motions of objects, enabling a manipulator to accurately throw them.
Similarly, \citet{gruenstein2021residualmodellearningmicrorobot} proposed learning residual actions for a simplified dynamics model for a legged microbot so that it transits to the same future states as a more complex dynamics model. In another study, \citet{sontakke2023residual} proposed learning a corrective external force policy to improve simulation accuracy for a buoyancy assisted legged robot.
Mentee Robotics has publicly stated that they applied RL to train a delta action model using real-world data
to overcome the sim-to-real gap on their humanoid, 
but the technical details of their approach remain unpublished
\cite{Mentee_Robotics_2024}. While we also apply RL to correct our simulation model, we specifically target the sim-to-real gap for the robot's actuators with harmonic drives, which are notably hard to model. This focus leverages the parts of the simulator that are more accurate (i.e., rigid body mechanics) to reduce overfitting and also avoids reliance on a motion capture system. 

\section{Conclusion} 
\label{sec:conclusion}
Legged manipulators promise enhanced strength and a larger workspace by coordinating arms and legs.
We proposed a training pipeline that first pre-trains a whole-body controller and then fine-tunes it using task rewards, while simultaneously reducing the sim-to-real gap via our UAN. Our experimental results on ball throwing, dumbbell lifting, and sled pulling demonstrate the viability of this approach. Future work may extend the sim-to-real calibration to additional subsystems and incorporate structural integrity constraints directly in the training.
Future work may extend the real-to-sim calibration to additional subsystems and incorporate structural integrity constraints directly in the training.

\section*{Acknowledgment}
We thank the members of the Improbable AI lab---especially Sandor Felber, Chen Bo Calvin Zhang, Srinath Mahankali, and Zhang-Wei Hong---for helpful discussions and feedback. We acknowledge Unitree Robotics for technical support provided for their robots. We are grateful to MIT Supercloud and the Lincoln Laboratory Supercomputing Center for providing HPC resources. We also acknowledge the MIT CSAIL Living Lab project for providing robot hardware. This research was partly supported by Hyundai Motor Company, the MIT-IBM Watson AI Lab, and the National Science Foundation under Cooperative Agreement PHY-2019786 (The NSF AI Institute for Artificial Intelligence and Fundamental Interactions, http://iaifi.org/) and the National Science Foundation Graduate Research Fellowship under Grant No. 2141064. This research was also sponsored by the United States Air Force Research Laboratory and the United States Air Force Artificial Intelligence Accelerator and was accomplished under Cooperative Agreement Number FA8750-19-2-1000. Research was sponsored by the Army Research Office and was accomplished under Grant Number W911NF-21-1-0328. The views and conclusions contained in this document are those of the authors and should not be interpreted as representing the official policies, either expressed or implied, of the United States Air Force or the U.S. Government. The U.S. Government is authorized to reproduce and distribute reprints for Government purposes, notwithstanding any copyright notation herein.

\section*{Author Contributions}
\small{
\begin{itemize}
\item \textbf{Nolan Fey} contributed to ideation, implementation of the entire system, experimental evaluation, and writing.
\item \textbf{Gabriel B. Margolis} contributed to ideation, implementation
of some parts of the system, and writing.
\item \textbf{Martin Peticco} contributed to hardware modifications and writing.
\item \textbf{Pulkit Agrawal} advised the project and contributed to its
development, experimental design, and writing.
\end{itemize}
}

\bibliographystyle{plainnat}
\bibliography{references}

\mbox{~}
\clearpage
\newpage

\appendices

\section{Training Details} \label{sec:appendix_training_details}

The PPO hyperparameters across all tasks are provided in \cref{tbl:ppo_hyperparameters}. The ranges for domain randomization are provided in \cref{tbl:domain_randomization}. \cref{tbl:wbc_rewards} details the WBC reward components, while Table \ref{tbl:uan_rewards} shows the reward function for UAN training.

\begin{table*}[htbp]
    \begin{center}
    \begin{small}
    \begin{sc}
    \begin{tabular}{lccc}
        \toprule
        Hyperparameter & UAN Value & Pre-train Value & Fine-tune Value \\
        \midrule
        Discount Factor & \underline{0.995} & 0.99 & 0.99 \\
        GAE Parameter & 0.95 & 0.95 & 0.95 \\
        Entropy Coefficient & 0.0 & \underline{0.01} & 0.0 \\
        Actor Learning Rate & Adaptive & Adaptive & Adaptive \\
        Critic Learning Rate & 5.e-4  & 5.e-4  & 5.e-4 \\
        KL Threshold & 0.01 & 0.01 & 0.01 \\
        Horizon & \underline{96} & 24 & 24 \\
        Number of Environments & 4096 & 4096 & 4096 \\
        Actor Minibatch Size & \underline{98304} & 24576 & 24576 \\
        Critic Minibatch Size & \underline{393216} & 98304 & 98304 \\
        \# of Mini Epochs & 5 & 5 & 5 \\
        Optimizer & AdamW & AdamW & AdamW \\
        Weight Decay & 0.01 & 0.01 & 0.01 \\
         \bottomrule
    \end{tabular}
    \end{sc}
    \end{small}
    \end{center}
    \caption{PPO Hyperparameters}
    \label{tbl:ppo_hyperparameters}
\end{table*}

\begin{table}[]
    \begin{center}
    \begin{small}
    \begin{sc}
    \begin{tabular}{lcc}
        \toprule
        Parameter & Min & Max \\
        \midrule
        Terrain Friction & 0.5 & 4.0 \\
        Terrain Restitution & 0.0 & 0.5 \\
        Terrain Roughness [\SI{}{\centi\meter}] & 0 & 2.5 \\
        Leg Joint Stiffness Scale & 0.9 & 1.1 \\
        Leg Joint Damping Scale & 0.5 & 1.5 \\
        Leg Stall Torque Scale & 0.9 & 1.1 \\
        Link Mass Scale & 0.5 & 1.5 \\
        Link Center of Mass Offsets [\SI{}{\centi\meter}] & -2 & 2 \\
        Encoder Offset [\SI{}{\radian}] & -0.05 & -0.05 \\
        Policy Lag Timesteps (\SI{5}{\milli\second}) & 0 & 6 \\
        \bottomrule
    \end{tabular}
    \end{sc}
    \end{small}
    \end{center}
    \caption{WBC \& Fine-tune Domain Randomization}
    \label{tbl:domain_randomization}
\end{table}

\begin{table*}[htbp]
    \renewcommand{\arraystretch}{1.5}
    \begin{center}
    \begin{small}
    \begin{sc}
    \begin{tabular}{lcc}
        \toprule
        Reward Component & Term & Scale \\
        \midrule
        EE Pose Tracking & $\sum_i^3 \frac{1}{3} {\rm exp}\left(-2 \abs{ p^{\rm cmd}_{i,t} - p_{i,t}} \right)$ & 5.0 \\
        Linear Velocity Tracking & ${\rm exp}\left(- 4 \abs{ \left[v^{\rm cmd}_{x,t}, v^{\rm cmd}_{y,t} \right] - \left[v_{x,t}, v_{y,t} \right] }^2_2 \right)$ & 2.0 \\
        Angular Velocity Tracking & ${\rm exp}\left(- 4 \left( \omega^{\rm cmd}_{z,t} - \omega_{z,t} \right) ^2 \right)$ & 1.0 \\
        Gait & $
         \sum_{i \in \{ \rm{FR, FL, RR, RL}\}} \{ \sim c_i\} ~\mathbbm{1}\{ p^i_{z,t}  < 0.043 \}
         $ & -0.5 \\
        No Slip & $
         \sum_{i \in \{ {\rm FR, FL, RR, RL} \}} \{ c_i \} ~ \exp \left( -0.1\abs{\boldsymbol{v}^i_t}^2_2 \right)
         $ & -0.5 \\
        Foot Clearance & $
         \sum_{i \in \{ {\rm FR, FL, RR, RL} \}} \{ \sim c_i \} ~ \abs{p^{{\rm cmd}, i}_{z,t} - p^i_{z,t}}^2_2 
         $ & -40.0 \\
        \hdashline
        Mechanical Power & $ \abs{ \boldsymbol{\tau}_{t} \cdot \dot{\boldsymbol{q}}_{t} }$ & -0.0001 \\
        Action Smoothness & $\abs{ \mathbf{a}_t - \mathbf{a}_{t-1} }^2_2 + \frac{1}{2} \abs{ \mathbf{a}_t - 2 \mathbf{a}_{t-1} +  \mathbf{a}_{t-2} }^2_2 $ & -0.05 \\
        Linear Velocity Z & $\abs{v_{z,t}}^2$ & -2.0 \\
        Angular Velocity XY & $\abs{\left[\omega_{x,t}, \omega_{y,t}\right]}^2_2$ & -0.05 \\
        Joint Positions & $\abs{\boldsymbol{q}^{\rm default}_t - \boldsymbol{q}_t}$ & -0.25 \\
        Collision & $ \{ \abs{ F^{\rm arm}_t }^2_2 > 0.1 \text{ or } \abs{ F^{\rm leg}_t }^2_2 > 0.1 \} $ & -5.0 \\
        Joint Position Limits & $\sum_i -\min(q_{i,t} - q^{\rm min}_{i,t},0) + \max(q_{i,t} - q^{\rm max}_{i,t},0) $ & -10.0 \\
        Contact Force & $\sum_{i \in \{ {\rm FR, FL, RR, RL} \}} \abs{ F^i_t}^2_2$ & -0.000004 \\
         \bottomrule
    \end{tabular}
    \end{sc}
    \end{small}
    \end{center}
    \caption{WBC Rewards}
    \label{tbl:wbc_rewards}
\end{table*}

\begin{table*}[htbp]
    \renewcommand{\arraystretch}{1.5}
    \begin{center}
    \begin{small}
    \begin{sc}
    \begin{tabular}{lcc}
        \toprule
        Reward Component & Term & Scale \\
        \midrule
        Joint Positions (L1) & $\abs{\boldsymbol{q}^{\rm real}_t - \boldsymbol{q}^{\rm sim}_t}$ & -1.5 \\
        Joint Positions (Relaxed) & ${\rm exp}\left(-100 \abs{\boldsymbol{q}^{\rm real}_t - \boldsymbol{q}^{\rm sim}_t}^2_2\right)$ & 4.0 \\
        Joint Positions (Moderate) & ${\rm exp}\left(-300 \abs{\boldsymbol{q}^{\rm real}_t - \boldsymbol{q}^{\rm sim}_t}^2_2\right)$ & 4.0 \\
        Joint Positions (Strict) & ${\rm exp}\left(-1000 \abs{\boldsymbol{q}^{\rm real}_t - \boldsymbol{q}^{\rm sim}_t}^2_2\right)$ & 5.0 \\
        Action Smoothness & ${\rm exp}\left( -0.5 \abs{\mathbf{a}_t - \mathbf{a}_{t-1} }\right) $ & 0.5 \\
         \bottomrule
    \end{tabular}
    \end{sc}
    \end{small}
    \end{center}
    \caption{Unsupervised Actuator Net Rewards}
    \label{tbl:uan_rewards}
\end{table*}

\section{Task Environments} \label{sec:appendix_task_environments}

The auxiliary rewards (and scales) for each task match those in \cref{tbl:wbc_rewards} underneath the dashed line.

\subsection{Ball Throwing}
The ball throwing task has three separate task states: \texttt{throw-set-up}, \texttt{throw}, and \texttt{settle}. The policy knows which state it is in by observing its task embedding, which includes a one-hot vector. Each state has a separate task reward for the desired behavior, such that 
\begin{align}
    r^{\rm ball}_t = b_{\texttt{throw-set-up}} &r^{\texttt{throw-set-up}}_t \nonumber  \\
    &+ b_{\texttt{throw}} r^{\texttt{throw}}_t + b_{\texttt{settle}} r^{\texttt{settle}}_t  \nonumber  
\end{align}
where $b_i$ is 1 when the environment is in state $i$ and 0 otherwise. The task reward for each state is
\begin{align}
    r^{\texttt{throw-set-up}}_t &= 5 \big(-(p_z - 0.6)^2 - g_x^2 - g_y^2 \nonumber \\ 
    & \quad\quad\quad\quad + ||q_{i,t} - q^{\rm ref}_{i,t}||_1 \big), \nonumber \\
    r^{\texttt{throw}}_t &=  20 \max \left( v^{\rm ball}_{x,t}, 0 \right) \nonumber \\  & \quad\quad\quad\quad + 20 \max \left(v^{\rm ball}_{z,t}, 0 \right)  
    - 0.5 \abs{v^{\rm ball}_{y,t}}^2, \nonumber \\
    r^{\texttt{settle}}_t &= 5 \big(-(p_z - 0.6)^2 - g_x^2 - g_y^2 \big). \nonumber 
\end{align}
The \texttt{throw-set-up} reward encourages an upright posture and tracking a trajectory to bring back the arm for a throw, then the \texttt{throw} reward is to maximize the ball's forward and upward velocities while minimizing its lateral velocity. The \texttt{settle} reward encourages the robot to stand upright after throwing and avoid falling over.

The environment transitions from \texttt{throw-set-up} to \texttt{throw} after \SI{2.5}{\second} in simulation time, and it transitions from \texttt{throw} to \texttt{settle} at \SI{3.5}{\second}. We set the ball's state to an arbitrary position at the start of the episode, and then we place it in the robot's bucket after \SI{1.5}{\second}. If the robot drops the ball in the \texttt{throw-set-up}, the environment terminates.

With this set up, we found that the policy would sometimes learn to lean during the \texttt{throw-set-up} task and eventually fall over unless it transitioned to the \texttt{throw} state. Thus, we modified the environment so that the policy remains in the \texttt{throw} state with a probability of $0.3$. Similarly, we added a probability that the robot enters the \texttt{settle} state on reset to help the policy learn to stay still after executing the throw.

We found that the policy may repeat the throwing motion multiple times during the \texttt{throw} state because it cannot sense whether it is holding the \SI{100}{\gram} ball through proprioception. Thus, we include a timer in the task embedding that linearly increments from 0 to 1 over 1 second during only the \texttt{throw} state, else it is 0. The policy can key into this timer to infer whether it has already thrown the ball. 

\subsection{Dumbbell Lifting}
We separate the dumbbell lifting task into two states: \texttt{snatch-set-up} and \texttt{snatch}. The task rewards for each state are
\begin{align}
    r_{\texttt{snatch-set-up}} &= 5\sum_i^3 \frac{1}{3} {\rm exp}\left(-2 \abs{ p^{\rm cmd}_{i,t} - p_{i,t}} \right) \nonumber \\ 
    & \quad \quad \quad\quad\quad - 0.005 ||F_t^{\rm EE} - F_{t-1}^{\rm EE}||, \nonumber \\
    r^{\texttt{snatch}}_t &= 4 p^{\rm EE}_{z,t} - 0.005 ||F_t^{\rm EE} - F_{t-1}^{\rm EE}|| \nonumber\\
    & \quad \quad \quad\quad\quad - \sum_i^{\{4,5,6\}} \abs{q^{\rm arm}_i}^2 . \nonumber 
\end{align}
The reward during \texttt{snatch-set-up} encourages tracking a reference that guides the EE near the ground. The force smoothness term penalizes the robot for suddenly slamming its EE into the ground. The \texttt{snatch} task reward is to maximize the end-effector height. To avoid damaging the robot's arm, we added a reward term penalizes deviations from the wrist's nominal, straightened position.

We used a similar strategoy to the throwing task to ensure stable transitions---the environment transitions from \texttt{snatch-set-up} to \texttt{snatch} after \SI{2.5}{\second} with a probability of $0.9$. Without this addition, the robot may fall over during an extended \texttt{snatch-set-up} state.

\subsection{Sled Pulling}
We model the sled as a virtual mass-spring-damper.
When an episode begins, the robot 
is at a distance $l$ from the sled.
The pulling force between the EE and the sled is defined as
\begin{equation}
    F^{\rm pull}_t = \max \left( 1000 \abs{d-l}_2^2 - \abs{\dot{d}}^2_2, 0 \right), \nonumber
\end{equation}
A lateral force is applied on the sled based on the friction and mass:
\begin{equation}
    F^{\rm sled} = F^{\rm pull}_x + \mu~m_{\rm sled}~g,  \nonumber
\end{equation}

The sled task also has two states, \texttt{sled-set-up} and \texttt{pull}, where the reward terms are
\begin{align}
    r^{\texttt{pull}}_t &= 5 \sum_i^3 \frac{1}{3} {\rm exp}\left(-2 \abs{ p^{\rm cmd}_{i,t} - p_{i,t}} \right), \nonumber \\
    r^{\texttt{pull}}_t &= -5 \abs{v^{\rm cmd}_{x,t} - v_{x,t} }^2_2 - \delta y_t - \delta \theta^{\rm yaw}_t - \sum_i^{\{4,5,6\}}q_i^2, \nonumber
\end{align}
where $\delta y_t$ and $\delta \theta^{\rm yaw}_t$ are deviations from the starting state.

The environment transitions from \texttt{sled-set-up} to \texttt{pull} after \SI{2.5}{\second} of simulation time with full probability. Every 7 seconds, forward base velocity commands are sampled  uniformly from 0 to $\SI{-1}{\meter\second^{-1}}$ with a probability of 0.8. Otherwise, the policy is given a command of $\SI{0}{\meter\second^{-1}}$. We found that the sled policies tended to drift excessively in the base-y and -yaw directions since the policy does not observe the robot's base position and orientation. Thus, when the forward base velocity command is non-zero, we set the lateral and yaw velocity commands as 
\begin{align}
    v^{\rm cmd}_{y,t} &= -0.5 \delta y_t, \\
    \omega^{\rm cmd}_{z,t} &= -0.5 \delta \theta^{\rm yaw}_t,
\end{align}
to help guide the policy to pull the sled straight. On hardware, we provide forward, lateral, and yaw velocity commands from a joystick.

\section{UAN Train \& Test Fits}
\begin{figure*}[htbp]
    \centering
        \centering
        \includegraphics[width=\textwidth]{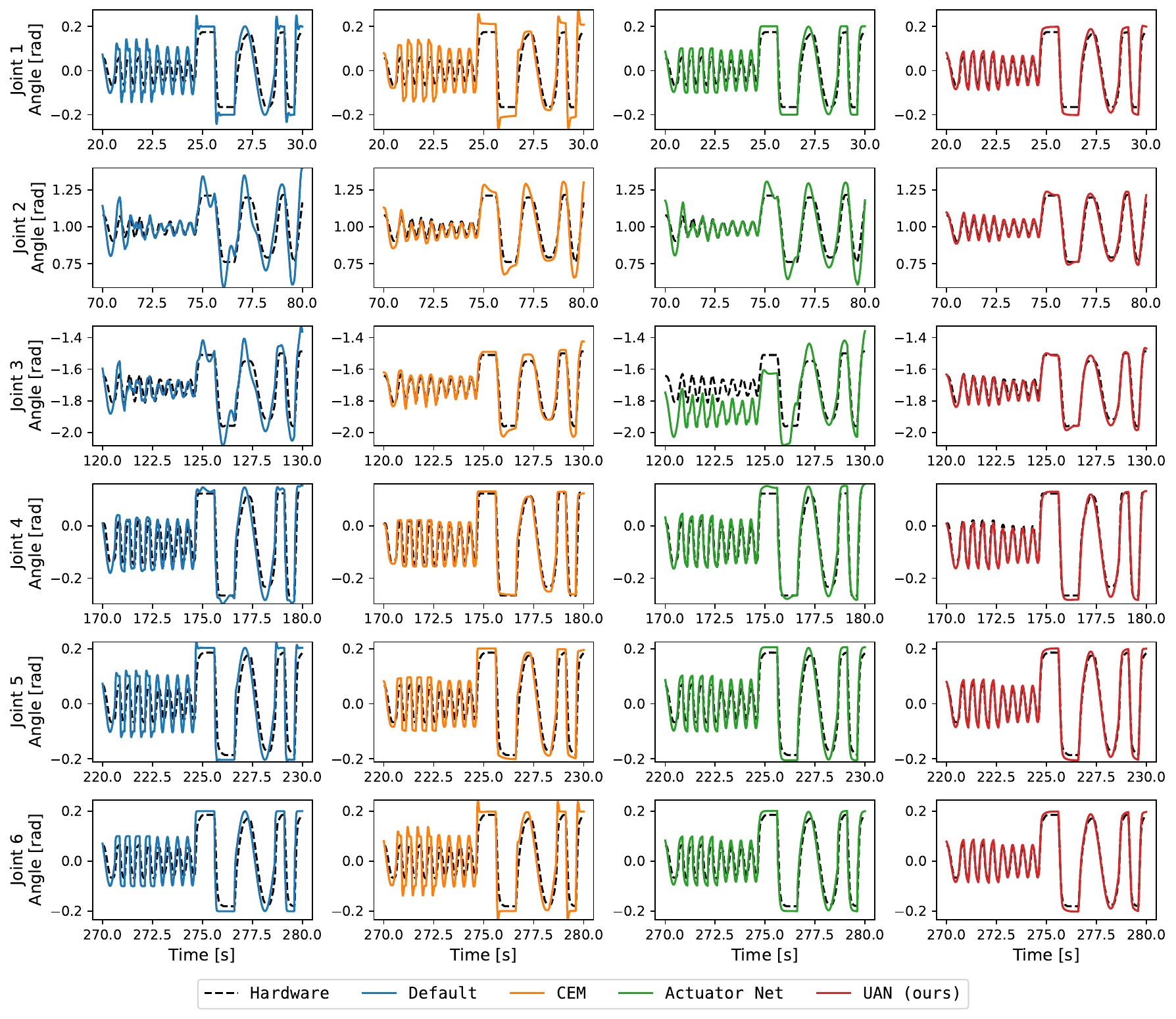}
        \caption{Comparison of system identification methods on open-loop rollouts of the training data.}
        \label{fig:sysid_comparison_train}
\end{figure*}

\begin{figure*}[htbp]
        \centering
        \includegraphics[width=\textwidth]{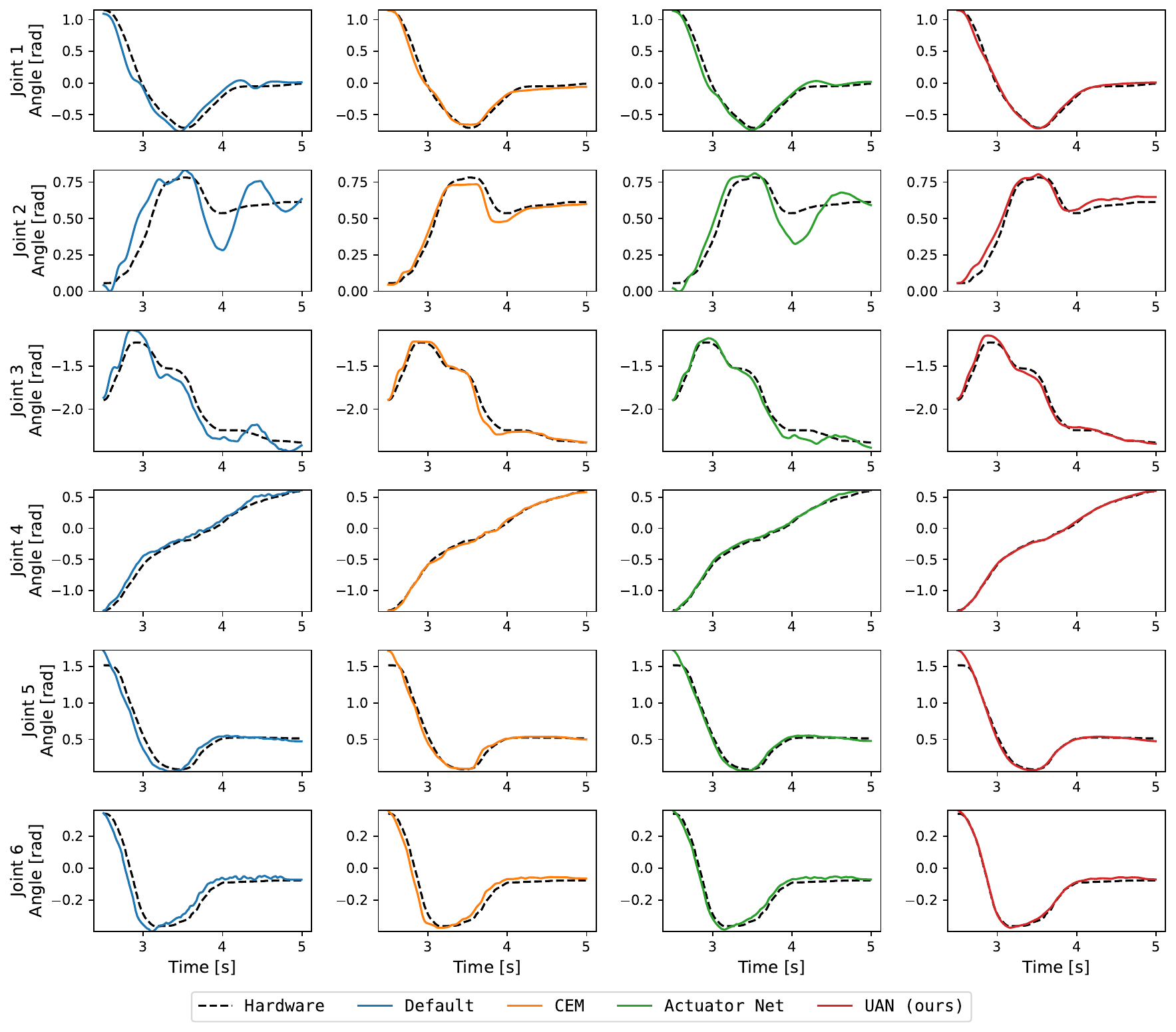}
        \caption{Sim-to-real gap for a throwing trajectory unseen during training.}
        \label{fig:sysid_comparison_test}
\end{figure*}

Simulator rollouts of each system identification method on training data and test data
are provided in Figures \ref{fig:sysid_comparison_train} and \ref{fig:sysid_comparison_test}.

\end{document}